# Can open source large language models be used for tumor documentation in Germany? - An evaluation on urological doctors' notes

*Stefan Lenz[1,4], Arsenij Ustjanzew[1], Marco Jeray[2], Meike Ressing[1,3], Torsten Panholzer[1]*

1. Institute of Medical Biostatistics, Epidemiology and Informatics, University Medical Centre of the Johannes Gutenberg-University Mainz, Mainz, Germany
2. Privacy, Compliance, and Risk Management Office, University Medical Centre of the Johannes Gutenberg-University Mainz, Mainz, Germany
3. German Childhood Cancer Registry (GCCR), Division of Childhood Cancer Epidemiology, Institute of Medical Biostatistics, Epidemiology and Informatics, University Medical Centre of the Johannes Gutenberg-University Mainz, Mainz, Germany
4. Corresponding author, e-mail: stefan.lenz@uni-mainz.de

May 9, 2025

## Abstract

**Background:** Tumor documentation in Germany is currently a largely manual process. It involves reading the textual patient documentation and filling in forms in dedicated databases to obtain structured data. Advances in information extraction techniques that build on large language models (LLMs) could have the potential for enhancing the efficiency and reliability of this process. Evaluating LLMs in the German medical domain, especially their ability to interpret specialized language, is essential to determine their suitability for the use in clinical documentation. Due to data protection regulations, only locally deployed open source LLMs are generally suitable for this application.

**Methods:** The evaluation employs eleven different open source LLMs with sizes ranging from 1 billion to 70 billion model parameters. Three basic tasks were selected as representative examples for the tumor documentation process: identifying tumor diagnoses, assigning ICD-10 codes, and extracting the date of first diagnosis. For evaluating the LLMs on these tasks, a dataset of annotated text snippets based on anonymized doctors' notes from urology was prepared. Different prompting strategies were used to investigate the effect of the number of examples in few-shot prompting and to explore the capabilities of the LLMs in general.

**Results:** The models Llama 3.1 8B, Mistral 7B, and Mistral NeMo 12 B performed comparably well in the tasks. Models with less extensive training data or having fewer than 7 billion parameters showed notably lower performance, while larger models did not display performance gains. Examples from a different medical domain than urology could also improve the outcome in few-shot prompting, which demonstrates the ability of LLMs to handle tasks needed for tumor documentation.

**Conclusions:** Open source LLMs show a strong potential for automating tumor documentation. Models from 7-12 billion parameters could offer an optimal balance between performance and resource efficiency. With tailored fine-tuning and well-designed prompting, these models might become important tools for clinical documentation in the future. The code for the evaluation is available from https://github.com/stefan-m-lenz/UroLlmEval. We also release the data set under https://huggingface.co/datasets/stefan-m-lenz/UroLlmEvalSet providing a valuable resource that addresses the shortage of authentic and easily accessible benchmarks in German-language medical NLP.

# Background

The goal of the tumor documentation in Germany is to capture tumor diagnoses, the treatment history and the corresponding outcome for all patients in a detailed and structured form. The compiled data should increase the transparency and quality in oncological care, improve treatment outcomes, and support research. All the data items necessary for describing the medical history of cancer patients are defined together with an overarching data structure in a unified dataset description, which is mandated by German federal law for adult cancer patients [1,2].

Within the framework of cancer registration, tumor diagnoses are to be coded using ICD-10 codes (International Classification of Diseases, 10th revision, [3]) to ensure consistency and comparability across registries. Key information such as the date of the first diagnosis, along with other clinically relevant attributes, must also be recorded. The documentation of first diagnosis dates is particularly important, as they form the basis for calculating cancer incidence rates in the population, a critical metric for public health monitoring and health services research.

Specialized tumor documentation units within hospitals are responsible for collecting and structuring the relevant information from medical records. Their primary role is to ensure that the data required for cancer registration is accurately extracted, coded, and prepared according to the mandated dataset specifications. The personnel in these units must read and interpret the medical documentation, which primarily comprises doctors' notes written in natural language. In doctors' notes, ICD-10 codes or codes from other medical classification systems are often not mentioned. Instead, the diagnoses are mainly described in text form, often using abbreviations and very heterogeneous wording. The presence of orthographic errors further complicates the situation. This way, coding information in a standardized way remains a predominantly manual task.

Large language models (LLMs), which are pre-trained on vast amounts of text data, have demonstrated significant success in a variety of text-based tasks across different domains. In areas like programming, LLMs shown impressive results in code generation [4,5] and fixing errors in code [6]. LLMs have also been highly effective at tasks such as summarization [7], translation, and answering complex questions [4].

The ability of LLMs to perform well in diverse, knowledge-intensive tasks suggests their potential for streamlining processes like medical coding and information extraction [8–10]. This new technological development is an opportunity for improving the tumor documentation process. It could significantly increase the accuracy and the amount of structured documentation as well as improve the timely availability of data, while reducing the workload for medical documentation staff. However, the application of LLMs to medical documentation in the German language is currently underexplored.

Foundational large language models, which are pre-trained on extensive text corpora, offer strong reasoning capabilities [11]. Very large LLMs such as GPT-4 exhibit a very strong performance across a variety of tasks [4]. Yet, the high level of data protection for healthcare data prohibits the upload of patient data to cloud-based models such as GPT-4 in clinical routine in Germany. While there is the possibility to use models such as ChatGPT legally with anonymized data for research purposes [12], the effort for anonymizing the documents would be too high to make this feasible in clinical routine. In Germany, each federal state (except for the common registry of Berlin and Brandenburg) maintains its own clinical cancer registry for adult cancer patients, and by law, only certain de-identified data may be shared with a central authority to avoid storing sensitive health information in a single location [13]. This also applies to data sharing with researchers or with other international institutions, which is regulated in the respective cancer registry acts of the federal states. Widespread use of, e.g., ChatGPT would undermine this principle of decentralization and data minimization, as it would involve transmitting sensitive personal health information of a large portion of citizens to a single US-based company.

Given the legal and regulatory background, only open source models that can be executed and fine-tuned locally actually have the potential to be adopted in practice for cancer registration. The models are required to have some knowledge of the German language as well, which restricts the number of possible foundational LLMs for this use case further. The question which of those open source models are capable enough to perform a task that is as complex as understanding the texts in doctors' notes, remains to be answered.

One example for this complexity in tumor documentation is the need to distinguish between definitive diagnoses of a patient, suspected or excluded tumor diagnoses, and diagnoses of relatives. In the case of a tumor diagnosis for the patient, the information about the treatment and the patient outcome needs to be linked to the tumor. Furthermore, the dates for diagnostics, disease progression and treatment are essential information for describing the patient history and the outcome. Therefore, we consider the extraction of the first diagnosis date as a representative example for the task of linking information in general and finding corresponding dates in particular. This is similar for many different data items in tumor documentation.

Here we focus on three tasks that are an integral part of the tumor documentation and also representative for others. These tasks are: finding the tumor diagnosis in text, coding the diagnosis using ICD-10, and determining when the tumor was first diagnosed. Based on a data set of urological doctors' notes from the university hospital in Mainz, we compare the performance of several open source models on these three exemplary tumor documentation tasks to assess the capabilities of current open source models with respect to tumor documentation.

# Methods

When writing this paper, we followed the TRIPOD-LLM statement [14] for reporting large language model research.

## Data set preparation

The initial data basis comprises 153 doctors' notes in the form of anonymized PDF documents. The primary purpose of the original data set was the use in a doctoral thesis evaluating GPT-4 for tumor documentation [15]. Regarding data protection, particular care was required when selecting the data for this purpose, as ultimately control over the data entered is lost when using Open AI's LLM. In order to avoid any data protection problems and violations of general personal rights, the data was selected and processed with the highest caution. The letters come from former patients with prostate cancer who have been deceased for at least 10 years in 2023. Before the doctors' letters were extracted from the clinical information system, names, addresses, dates of birth, and other personal identifying data were removed. These files were then exported from the hospital information system in form of PDFs and redacted to ensure the anonymity of the letters. Therefore, further data was removed, such as the exact date of admission or examination, as well as information that could allow conclusions to be drawn about genetic diseases, family members or third parties.

Limiting the input to only the relevant parts of the PDF documents reduces the time for querying the LLMs substantially. In addition to the computational costs for using the complete documents as inputs, there is also the problem of degradation if relevant information is embedded in very long inputs [16]. Therefore, we added a preprocessing step to extract only the relevant sections of the documents that contain a synopsis of the patient's medical history.

The text was extracted from the PDFs with the information about the positions of the text lines on the pages. Based on this positional information, we clustered the text into text blocks and identified the headings of the blocks. Subsequently, only the blocks with the German word for diagnosis in the header were used further. This process was sufficient to identify almost all text blocks that summarized the relevant diagnoses and treatments in the documents. These text snippets were put into a Microsoft Excel document. Initially, a single annotator added ICD-10 codes for the tumor diagnoses and the corresponding first diagnosis dates in a different column. The Excel document was then transformed into an XML file that served as the primary basis for the experiments [9].

To ensure the annotation quality, the annotation process was repeated by three different independent annotators. The agreement between the four annotators with respect to ICD-10 coding and first diagnosis date attribution was calculated using Fleiss' kappa and Krippendorff's alpha. (The two statistics were computed using the Python packages "statsmodels" [17] and "krippendorff" [18], respectively, and had identical values when rounding to two digits.) For the binary variable of having a tumor diagnosis present in a text snippet, the inter-rater agreement reached 0.90. To evaluate agreement at the level of ICD-10 codes, we transformed the code sets into a categorical variable, with each unique ICD-10 code set annotated by any annotator treated as a distinct category. This yielded a Krippendorff's alpha of 0.85. This value can be seen as a lower bound of the true agreement, as the disagreement expected by chance is underestimated due to being calculated for the smaller set of observed combinations instead of the full space of possible ICD-10 code sets. For the agreement on first diagnosis dates at the level of individual diagnoses, Krippendorff's Alpha was 0.84. Overall, these values indicate a very high, almost perfect, inter-rater agreement [19].

The cases with differing annotations were analyzed and discussed, which resulted in a limited amount of changes to the original annotation [9]. While most adjustments concerned only the assigned codes or diagnosis dates, in one case we extended the text snippet with additional context from the original document to enable more precise ICD-10 coding, as the tumor dignity had previously not been derivable from the original snippet alone.

A clear consensus could be reached in all cases for the diagnoses and ICD-10 codes. Attributing the first diagnosis dates proved more difficult. The first diagnosis date is defined as the date when the tumor diagnosis was first established by a physician, based either on clinical evaluation or microscopic confirmation [1]. For the annotation purposes here, we used all dates that were clearly attributable directly to the tumor in the text and not to treatments. In many cases, this can be determined very clearly, as the dates are marked with "ED", which is the abbreviation of "Erstdiagnose" (German for "first diagnosis"). However, in some cases, it was difficult to tell whether a date referred only to the treatment or also to the diagnosis. Examples for this include phrases like "Z. n. Prostatektomie bei Prostata Ca 2008" (approximate translation: history of prostatectomy for prostate cancer 2008) or "Prostata-Karzinom Gleason 6, Zustand nach transurethraler Resektion 2008" (approximate translation: prostate cancer Gleason 6, history of transurethral resection 2008). These cases were harmonized by annotating only those cases where a clear connection of the date to the tumor diagnosis was apparent for all annotators. This reflects the instruction given in the prompts to only use dates clearly linked to the first diagnosis.

For the publication of the data, the consented data set was finally transformed into the standard dataset format on HuggingFace using the Python "Datasets" package.

The product of the text extraction and annotation process was a set of 149 annotated text snippets. The total number of diagnoses in the collection of text snippets is 157. Of those tumor diagnoses, 82 have a first diagnosis date annotated. The dataset comprises 25 patients in total, with some overlap in terms of content between snippets derived from the same patient history. The distribution of the text snippets and the different types of tumor diagnoses can be found in Table 1. A detailed mapping of the relationship between patients, text snippets, and diagnoses is provided in Table S1 in the supplementary material.

Table 1 Distribution of tumor diagnoses in the annotated text snippets. The proportions reported in the last column are all relative to the total number of snippets (149).

| Text snippets containing... | Number of snippets | Proportion of snippets |
|---|---|---|
| at least one tumor diagnosis | 112 | 75% |
| multiple tumor diagnoses | 26 | 17% |
| no tumor diagnosis | 37 | 25% |
| C61 (prostate cancer) | 103 | 69% |
| C67 (bladder cancer) | 14 | 9.4% |
| C34 (lung cancer) | 10 | 6.7% |
| C83 (B-cell lymphoma) | 9 | 6.0% |
| C09 (tonsil cancer) | 9 | 6.0% |
| C45 (malignant mesothelioma) | 4 | 2.7% |
| C90 (multiple myeloma) | 4 | 2.7% |
| C68 (urinary tract cancer, unspecified) | 2 | 1.3% |
| C18 (colon cancer) | 1 | 0.67% |
| D47 (Monoclonal gammopathy) | 1 | 0.67% |

ICD-10 codes for the diagnoses do not appear in the texts in most cases. In one of the coded diagnoses, a wrong ICD-10 code (C83.3 instead of C82.2 for coding a "Large-cell B-cell Non-Hodgkin-Lymphoma") was used in the original text. This was corrected in the text because the purpose of the evaluation is the exploration of the coding capabilities with respect to the correct and current ICD-10 catalogue.

To further strengthen the anonymization in the final data set, the dates in the texts were altered: For each snippet, all the dates therein were shifted by a random amount of time that is constant in the snippet. Thereby, the time spans of events in each text snippet remain constant and realistic, while a hypothetical use of the dates for patient reidentification is prevented.

Except for the modifications described above, the dataset was retained in the form that was returned by the extraction mechanism during conversion from PDF to text. The text extraction was performed using the software "PyMuPdf" [20]. The original documents were created digitally and are not scans from paper documents. Therefore, image processing techniques and Optical Character Recognition (OCR) were not necessary and errors introduced by OCR can be ruled out in the data. Further cleaning was not performed after the text extraction. In particular, special characters have not been removed and no additional manual formatting for clarification was performed. This way, the dataset remains a realistic real-world output from a PDF text extraction process.

## Model selection

Due to data protection constraints, only models that can be executed locally can be used to improve the tumor documentation process in German hospitals. Therefore, our evaluation focused solely on open-source models, as these are the only practically relevant options. In this study we define open source models as those models with freely available weights and a license allowing modification and commercial use. We also considered only models that can be run on our local infrastructure at a reasonable speed. Our evaluation in particular was constrained by the maximum available infrastructure, which consisted of a server with three NVIDIA A40 GPUs, each with 48 GB VRAM. These requirements excluded very large open source models such as Command R+ [21] or Llama 3.1 405B [22].

For using LLMs for the tumor documentation in Germany, it is necessary that the LLMs have been trained on German texts. In this study, Mistral 7B [23] from Mistral AI and Llama 3.1 8B [22] from Facebook/Meta were utilized as popular open source LLMs that have been trained on a large corpus of data, including German texts.

We also wanted to examine variants of these models that have received additional training with German texts or with texts from the medical domain to see whether they can improve the results in the experiment. We included BioMistral [24] as a variant of Mistral 7B that has been fine-tuned on texts from the medical domain. We furthermore included LeoLM [25], which is a model trained on basis of Llama 2 [26] with more German text.

The company VAGOsolutions, a software company specializing in LLMs, uses a proprietary dataset for additionally training LLMs with the intention to improve the performance on German text [27]. VAGOsolutions brands these models with "SauerkrautLM". We included their Llama 3.1 8B variant to see whether we can get an improvement compared to the base model in this scenario with German clinical documentation.

Besides the mentioned models, which have sizes of seven or eight billion parameters, we also evaluated both larger and smaller models. This includes the 12-billion-parameter Mistral NeMo [28] as well as two larger open-source models Mixtral 8x7B [29] and Llama 3.1 70B [22]. As smaller models, EuroLLM 1.7B [30], Llama 3.2 1B [31], and Llama 3.2 3B [32] were examined.

Our evaluation used the open source models as they are provided. No fine-tuning was performed on the models. We restricted this evaluation to LLMs that can be directly prompted with textual input and do not require additional fine-tuning on downstream tasks, as is the case with models building on the BERT architecture [33]. Thereby we are assessing the baseline performance of the models, providing a practical reference point for future improvements through task-specific training or adaptation. For all models, the instruction-tuned variants, if available, were used, as the instruction tuning makes the models better at adhering to instructions given in the prompt [34].

Table 2 provides an overview of the key properties of each model, including their sizes, the total volume of their training data and the extent of German text coverage. For all models examined in this study, the training data has not been made publicly available, unlike the model weights, which are open source. Information regarding the composition of the training data is generally sparse. We were unable to find any specific information regarding the proportion of medical texts, or particularly German medical texts, in the training data of these models.

Table 2 Overview of evaluated models, their sizes and training data volumes. The models are sorted by size, defined as the number of learned parameters (weights and biases), as specified in their respective model cards on HuggingFace. "ND" indicates entries where the amount of training data has not been disclosed.

| Model | No. of parameters / $10^9$ | Amount of training data (No. of tokens / $10^9$) | Amount of German training data (No. of tokens / $10^9$) | Official German language support | Model card |
|---|---|---|---|---|---|
| Llama 3.2 1B | 1.23 | 9,000 | ND | ✓ | [31] |
| EuroLLM 1.7B | 1.66 | 4,000 | 240 | ✓ | [35] |
| Llama 3.2 3B | 3.21 | 9,000 | ND | ✓ | [32] |
| LeoLM 7B Chat | 6.74 | 2,000 + 68 | 68 | ✓ | [25] |
| BioMistral | 7.24 | ND | ND | ✗ | [36] |
| Mistral 7B v0.3 | 7.25 | ND | ND | ✗ | [37] |
| Llama 3.1 8B | 8.03 | 15,000 | (x) < 1,200 | ✓ | [38] |
| Llama 3.1 SauerkrautLM 8B | 8.03 | 15,000 | (y) < 1,200 | ✓ | [39] |
| Mistral NeMo | 12.2 | ND | ND | ✓ | [28] |
| Mixtral 8x7B | 46.7 | ND | ND | ✓ | [40] |
| Llama 3.1 70B | 70.6 | 15,000 | (x) < 1,200 | ✓ | [41] |

(x) Less than 8% of the total training data is multilingual and German is one of 7 officially supported non-English languages [22].
(y) The exact amount of training data for this model has not been stated on the model card but the SauerkrautLM dataset used by VAGOsolutions for fine tuning similar models has about 70,000 tokens [42].

Prompt design

There are three steps in the evaluation. Each of the steps uses different prompts for obtaining structured information from the text snippet and the output from previous steps. The core parts of the prompts and the corresponding objectives are shown in Table 3. The original prompts are in German and contain more detailed explanations. These complete prompts can be found in Appendix B.

The core parts of the prompts that are shared among different prompt variants explain only general concepts that are relevant for the task. In particular, they explain the concept "tumor diagnosis" (Step 1) in the context of tumor documentation and what counts as a "first diagnosis date" (Step 3) here. The core prompts were not designed for detecting tumor diagnoses specific to urology and contain no instructions specific to this domain. Thereby, we aimed to measure the performance of LLMs for detecting all sorts of tumor diagnoses instead of detecting specific ones.

Table 3 Overview of the three steps of the evaluation with their objectives and the core prompt phrasing translated from German to English. In Step 3, two variants have been explored. Variant 3a asks directly for the first diagnosis date of a given tumor diagnosis. Variant 3b first uses a rule-based algorithm to identify possible dates in the proximity of the diagnosis labels and then tries to verify a single date by asking with one prompt per identified date.

| Step | Objective | Core prompt |
| --- | --- | --- |
| 1 | Find labels of tumor diagnoses of the patient in the larger snippets | "Are there one or more tumor diagnoses in the following diagnostic text? [...] Respond with a JSON array that includes the tumor diagnoses as strings, or with an empty array if there are no clear tumor diagnoses." |
| 2 | Find the ICD-10 codes for the labels of tumor diagnoses | "What is the ICD-10 code for the diagnosis [DIAGNOSIS]? Respond briefly with the 3-character ICD-10 code only." |
| 3a (One prompt per diagnosis) | Find first diagnosis dates for the ICD-10 coded diagnoses by asking for the date directly in a single prompt for each diagnosis | "In the diagnostic text, you have identified the diagnosis [DIAGNOSIS]. [...] Respond with the date if it describes the time of the first diagnosis of the tumor [DIAGNOSIS], or with 'No'." |
| 3b (One prompt per possible date) | Find first diagnosis dates for the ICD-10 coded diagnoses by asking for verification of possible dates in the text | "In the diagnostic text, you have identified the diagnosis [DIAGNOSIS]. Furthermore, the following date is found in the diagnostic text: [DATE]. Is this the date of the initial diagnosis of the tumor disease [DIAGNOSIS]? Respond with 'Yes' or 'No'." |

We tried different styles of prompting to refine the prompts for investigating the effect of additional information in the prompt. For this purpose, we tried prompting variants that give the models information specific to the domain of urology, while other prompts give similarly structured information that is not concerned with urology but has examples from a fictitious gynecological unit. We also explore the effect of the number of examples used in few-shot prompting.

The urological examples in Step 1 are not directly taken from the dataset but constructed to reflect the main tumor diagnoses from the dataset, i.e., "prostate carcinoma" and "urothelial carcinoma of the bladder". The fictitious examples from a gynecological unit in Step 1 concern "breast cancer" and "ovarian cancer", which do not occur in the text snippets since these only concern male patients. Positive examples (with a tumor diagnosis) and negative examples (no tumor diagnoses) are presented in equal numbers. In the sequence of examples, there are always negative examples that have a suspected diagnosis or an exclusion of the mentioned tumor diagnoses. The intention is to ensure that the model not only understands the terms of the tumor diagnoses but also the context in which these terms are used.

In Step 1, the models are asked to return a JSON array with all the tumor diagnoses in the text snippet as strings. If the JSON array is empty, there are no tumor diagnoses in the text. To evaluate the ability of the models to find tumor diagnoses in the texts, we examine the performance measures for the test whether a tumor diagnosis occurs in the text snippet or not. The test considers the presence of data in the JSON array as the predicted result. The predicted result is compared to the presence of annotated tumor diagnoses as the ground truth.

The accuracy of this test is used to determine the best model for extracting the tumor diagnoses. Non-interpretable values are counted as wrong results for this purpose since a good model should be reliable with respect to correctness and usability of the answers. In addition to the accuracy, we use the sensitivity and specificity to further investigate whether the model is able to understand and apply the concept of a tumor diagnosis as used in the tumor documentation process.

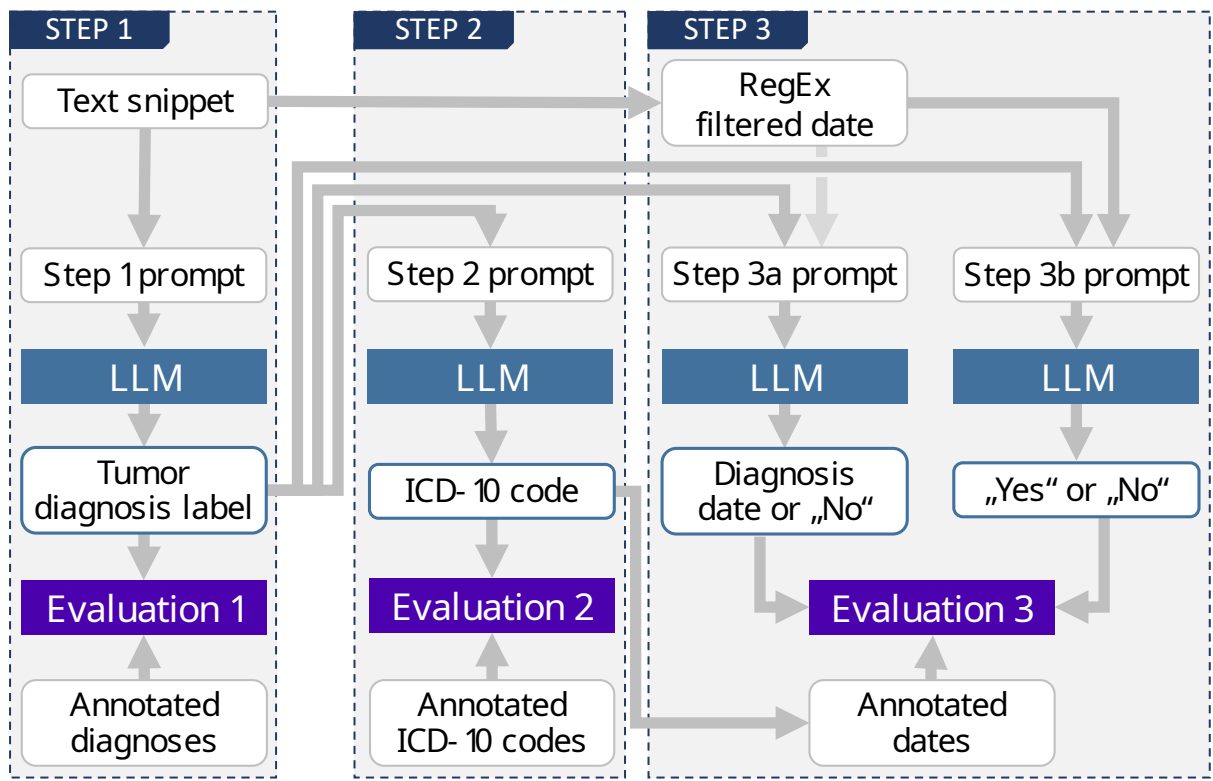

*Figure 1 Dataflow diagram illustrating the inputs and outputs of the three steps of the evaluation process. Some prompting variants in Step 3a included dates filtered by regular expressions (RegEx), which is symbolized with the light-gray arrow. White boxes illustrate data elements, while the LLM and evaluation boxes describe processes.*

Steps 2 and 3 of the evaluation each require the results from the previous steps. For Step 2 this is obvious when looking at Table 3 as the prompt directly uses the diagnosis labels to find the ICD-10 code. In Step 3, the resulting ICD-codes from Step 2 are needed to identify the diagnosis in the annotated data. Identifying the diagnosis in the data is then in turn necessary to find the corresponding date that is needed for the comparison with the model answer.

In all three steps, the model answers in the form of text, which needs to be transformed into values that can be processed further. The prompts are designed with the intention to obtain easily parsable answers from the models. In Step 1, the answer is supposed to be a textual representation of a JSON array of strings, in Step 2 it should be an ICD-10 code, and in Step 3 either a date in German format or Yes/No (in German). Answers that did not conform to the requested format, and which could therefore not be parsed, were classified as unusable values ("NA") in the evaluation.

The prompts were not systematically optimized in the sense of using programmatic methods, controlled experiments, or automated search procedures. They were primarily designed with the intention to convey the task as clearly as possible in human language. Based on preliminary results, we made a few targeted manual adjustments to address common biases observed in the model answers. In Step 1, the models tended to have a low specificity, i.e. they reported various facts as tumor diagnoses that did not constitute an actual tumor diagnosis. The prompts were adapted to mitigate this: We explicitly instructed the models in the prompt not to report therapies, symptoms or other diseases as tumor diagnoses and to be extra cautious and thoughtful about reporting something as a tumor diagnosis.

In Step 2, we also used preliminary results to inform the prompt design. We observed the problem that the ICD-10 code C78 in particular was frequently returned as answer by some models. While the answers using the code were technically correct in most observed cases, these answers are not desired answer because C78 is used to code metastases, not the primary tumor. Therefore, we instructed the models in the prompt not to use the codes C77, C78, and C79, which are all codes for secondary neoplasms and should not be used for documenting the primary tumor diagnosis.

To minimize further bias introduced through prompt design and to avoid skewing model performance in a particular direction, we took several deliberate measures across all steps of our evaluation. In Step 1, we ensured that few-shot prompts included both positive examples (containing tumor diagnoses) and negative examples (containing none), resulting in a balanced number of examples (0, 2, 4, or 6). This was done to prevent the model from being implicitly guided toward higher sensitivity by exposure to only positive cases. Balancing positive and negative cases is important because language models may exhibit what is referred to as "majority label bias" [43], where the predominance of one class in the examples can systematically influence the model predictions in that direction. Additionally, the pattern of negative (N) and positive examples (P) in Step 1 was not easily predictable (PN, PNNP, and PNNPNP for two-, four- and six-shot prompting, respectively) to avoid that the models pick up the order and answer accordingly. The analysis of results across different prompt types enables the assessment of a potential "recency bias", which may occur if models are disproportionately influenced by the final examples in few-shot prompting [43]. To be able to examine domain-label bias [44] introduced by using domain-specific words and to examine how well the model answers generalize beyond the context given in the examples, the few-shot examples in Steps 1 and 2 included cases from both urology and gynecology. This allowed us to contrast prompts of varying informativeness, including those that may offer helpful guidance but risk being overly suggestive. In Step 3, where the model was asked to identify the first diagnosis date or state the absence of such a date, the few-shot prompts included three examples: two with valid dates (one month-based and one year-based) and one negative example (no date). This mix ensured the model was exposed to different date formats while still being cautioned against overprediction. In prompt variants where a date was preselected and the model was asked whether it represented the first diagnosis date, we similarly included both month-based and year-based examples, balanced with negative examples, which lead to four examples in total. Across all these designs, we maintained symmetry in example composition to both mitigate bias in model predictions and to allow for the assessment of potential sources of bias.

For finding the first diagnoses dates in Step 3, we added a rule-based baseline approach to contrast it with the results from prompting the LLMs. This heuristic uses the Levenshtein distance for finding the text part with the least Levenshtein distance to the tumor diagnoses label returned from Step 1. Dates can easily be identified via regular expressions in the text. Dates that have the least character distance to the tumor diagnosis label are most likely to belong to the first diagnosis of the tumor. We explored variants of this rule-based approach by looking at the same line of the tumor diagnosis label, and up to two lines distance, for identifying first diagnosis dates. Some variants of prompting in Step 3 also include dates filtered by regular expressions regardless of the distance to the diagnosis label. These variants should help to explore whether the explicit mentioning of dates in the prompt can aid the LLMs in finding the correct date.

When applying LLMs for a documentation task in practice, it would be possible to utilize different models for different sub-tasks to combine their strengths. To account for such a scenario, we ran the evaluation for Step 2 and Step 3 two times: one time with the output from the same model and the prompt that worked best with this model as input, and one time with the output from the best model/prompt combination overall as input. This allows an overall comparison of the models as well as a task-specific one.

### Software implementation

The complete code for running the evaluation is published on GitHub (https://github.com/stefan-m-lenz/UroLlmEval) and available under the MIT license. The data is released on HuggingFace (https://huggingface.co/datasets/stefan-m-lenz/UroLlmEvalSet) to serve as a benchmark data set that can also be used independently of this evaluation.

The code for this evaluation is written in the Python language, building on the Python "transformers" package [45] for executing the models locally and the "pandas" package for analyzing and organizing the data [46]. A value of zero was used for the temperature in the LLM inference algorithms, which limits the diversity of the answers of the LLMs and allows the results to be reproduced. The entire evaluation can be run by executing one Python script that

- 12 -

ties the code for the different steps and for the subsequent analyses of the results together. The code was run on a machine with three NVIDIA A40 GPUs, each with 48 GB of VRAM, for the experiments. The larger models Mistral NeMo 12B, Mixtral 8x7B and Llama 3.1 70B were quantized using 8-bit quantization to accelerate computation and to fit within the GPU memory constraints [47]. Running the complete evaluation took approximately 25 hours. Running only the nine smaller models that fit into one of the GPUs took around 10 hours.

# Results

## Results for Step 1: Detecting tumor diagnoses

In Step 1, the models return a JSON array of tumor diagnoses extracted from each text snippet, with an empty array indicating no diagnosis. We evaluate model performance based on accuracy, sensitivity, and specificity by comparing the presence of predicted diagnoses to the annotated ground truth.

As can be seen in Figure 2, Llama 3.1 8B is in Step 1 the overall best model in terms of accuracy. This means that the largest models Mixtral 8x7B and Llama 3.1 70B are surprisingly not also the best models in this scenario. The SauerkrautLM variant with a focus on the German language is very close to the Llama 3.1 8B model but does not improve upon the base model. Mistral NeMo 12B performs also very similar to Llama 3.1 8B in this test.

LeoLM and BioMistral perform clearly worse than the other models with at least 7 billion parameters. They also produce more values that cannot be parsed, which indicates that they have more trouble understanding the instructions and following them closely. The two smallest models, EuroLLM 1.7B and Llama 3.2 1B, are furthermore clearly weaker than all the other, larger models. Llama 3.2 3B shows a performance that is very close to the larger model Mistral 7B.

In Figure 2 all the values for the different prompting variants are shown. The variance of the performance measures across different prompt variants allows drawing conclusions about the models' general capabilities. The more knowledge the models possess, the less dependent they are on additional instructions, making it easier for them to interpret explanations and examples. There is less impact of the prompting variants on the performance of the larger models. EuroLLM 1.7B and LeoLM 7B Chat show the most variance in the performances of the prompting variants. The impact of foundational knowledge and general comprehension of instructions is clearly reflected in both the performance and its stability in these results.

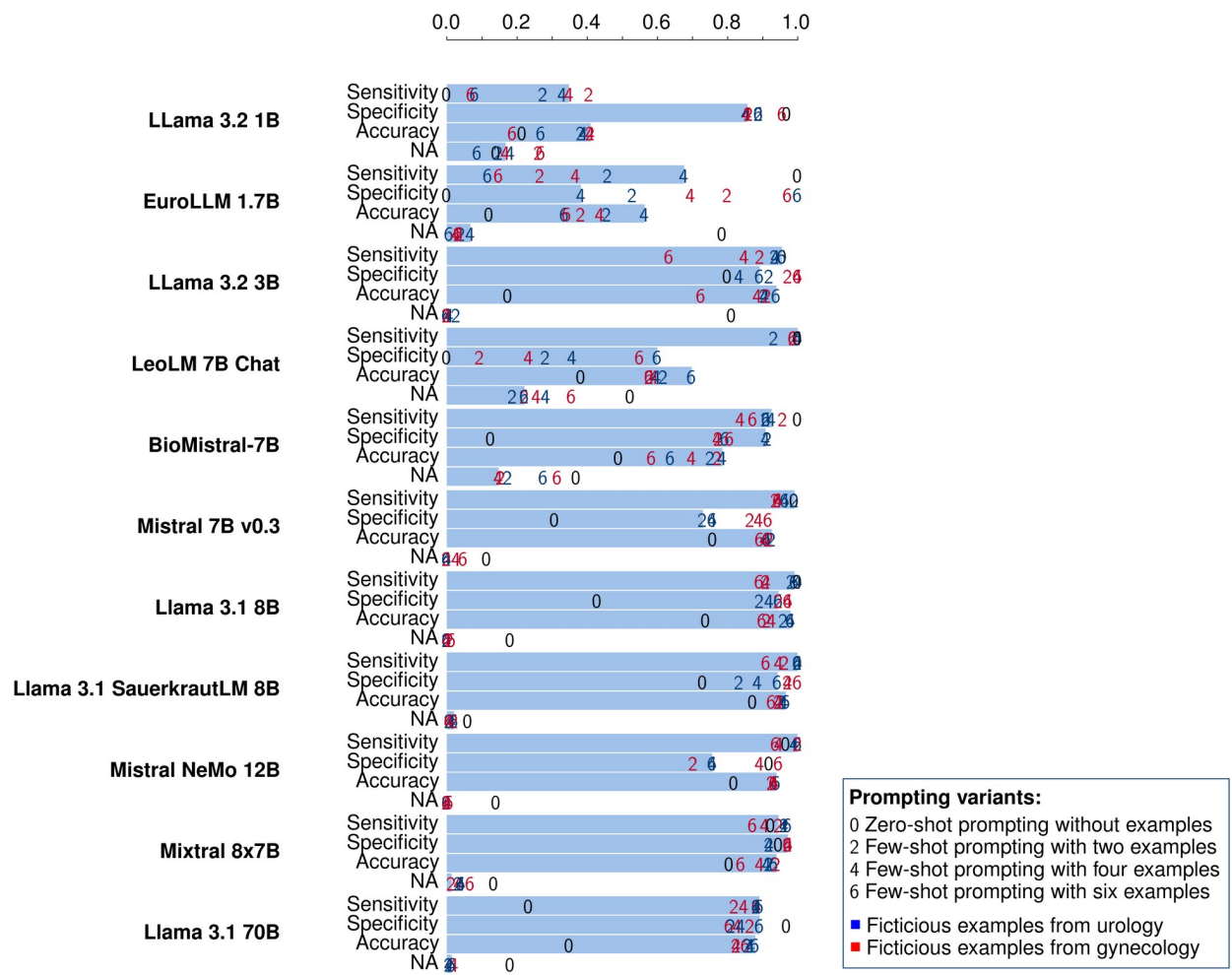

Figure 2 Sensitivity, specificity and accuracy for detecting a tumor diagnosis in a text snippet using different models. The models are sorted by size, starting the with the smallest model (1 billion parameters) and ending with the largest model (70 billion parameters). The values for different prompt types are marked with colored numbers (see legend). The bars indicate the best value/prompt type with respect to the accuracy for the given model. The number of not usable values returned by the LLM is indicated in the bar labeled with "NA". The sensitivity and specificity is calculated only with the usable values. An interactive visualization of the data, designed to facilitate selective exploration, is available as Figure S1 in the supplementary materials.

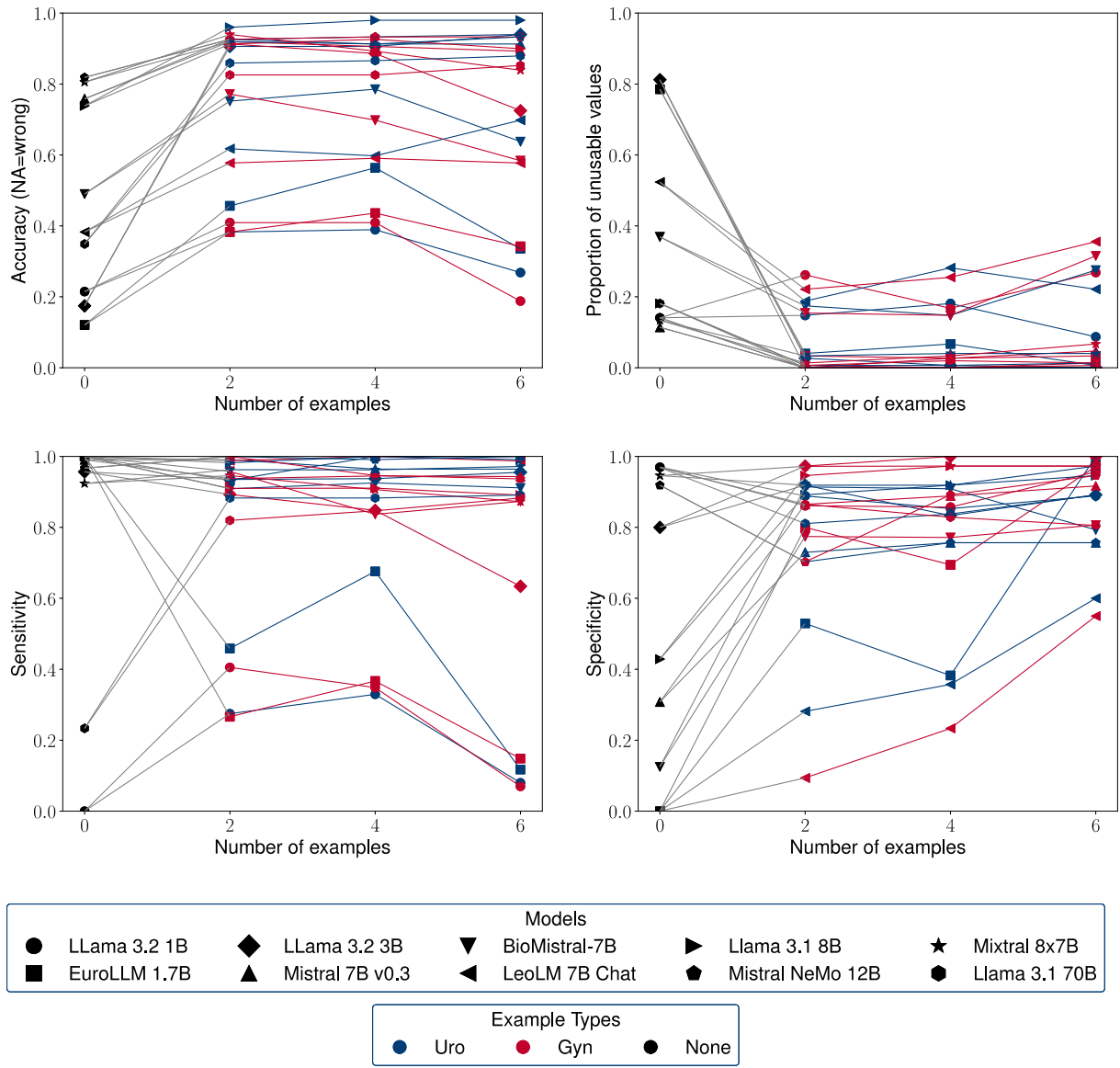

*Figure 3 Influence of the number of examples used in few-shot prompting for detecting tumor diagnoses on the performance of the different LLMs. The values displayed here are the same as in Figure 2. The plot shows the number of examples on the x-axes and the different performance metrics on the y-axis to highlight the effect of increasing the number of examples in few-shot prompting. Additionally, the effect of the example type (fictitious examples from a urological (Uro) vs. a gynecological (Gyn) unit) can be observed. (The model Llama 3.1 SauerkrautLM 8B with a trajectory close to Llama 3.1 8B has been left out to make the figure more readable.)*

Figure 3 shows the influence of the number of examples used in few-shot prompting on the model performance. Most models profit from a small number of examples. They also profit from examples that come from a different medical domain such as the examples from the fictitious gynecological unit. This shows that the examples do not only help the LLMs to simply recognize the names of the relevant tumor diagnoses. LLMs can also learn and apply abstract concepts from few-shot prompting like suspected diagnoses, excluded diagnoses, and the difference of diagnoses and therapies.

The models with at least 7 billion parameters behave in a similar way when increasing the number of examples. The relationship between the number of examples and the performance of smaller models exhibits less predictable patterns.

Presenting examples leads to fewer non-interpretable values. The zero-shot variants of the prompts yield a substantially higher proportion of unusable outputs than all the few-shot prompting variants. This is also true for the larger models. Another effect of adding examples to the prompts is the increase in the specificity, which can be observed in almost all combinations of example types and models. The effect of the number of examples on the specificity is independent of the effect on the proportion of non-interpretable values, as the NA values are ignored when calculating the specificity.

The observed increase in specificity with more few-shot examples across most models indicates a lack of susceptibility to positional bias stemming from the order of the examples. Instead, many models are rather skewed in the direction of over-reporting diagnoses as tumor diagnoses. This can be seen in lower values for the specificity compared to the sensitivity in the zero-shot prompts, which is observed for most models. This happens although the formulation of the zero-shot prompt explicitly emphasizes to only report tumor diagnoses and not any other diagnoses.

The examples have little impact on the sensitivity of the models of at least 7 billion parameters. Counterintuitively, increasing the number of examples even reduces the sensitivity of the smaller models EuroLLM 1.7B, Llama 3.2 1B, and Llama 3.2 3B. This suggests that these smaller models lack a solid understanding of the basic instructions and become confused about the concept of what constitutes a tumor diagnosis.

As there is little effect of more examples on the sensitivity, the improvement in the accuracy can be attributed entirely to the increased specificity. This shows that presenting negative examples together with positive ones is a suitable strategy not only for improving the specificity but for the overall accuracy as well, at least for the models having 7 billion parameters or more.

### Results for Step 2: ICD-10 coding of tumor diagnoses

We find the task-specific comparison of the models more intriguing than evaluating their performance across all tasks individually. Figure 4 shows the results from the translation of the diagnosis labels from Step 1 to ICD-10 codes. There we used the output from the best model/prompt combination from Step 1 for all models because we want to assess the model performance directly on the ICD-10 coding without the effect of the diagnosis label extraction.

Appendix A contains the tables with all the results, both for the task-specific and the overall performance. The results using the best/model prompt combination from the previous step are better than the ones where the same model is employed in all steps.

In Step 2, we compare the models using three different metrics: Analogously to the sensitivity, we want to know the proportion of snippets where all annotated diagnoses are found by the model ("All diagnoses found" in Figure 4). Similarly to the specificity, we report the proportion of snippets where the model answers only with correct ICD-10 codes ("No incorrect diagnosis"). The proportion of snippets with no incorrect diagnoses can also be seen as a measure for the absence of hallucination in this task. To identify the best model in this task, we use the proportion of snippets where all ICD-10 codes have been identified and no incorrect diagnoses have been returned ("Snippet correct").

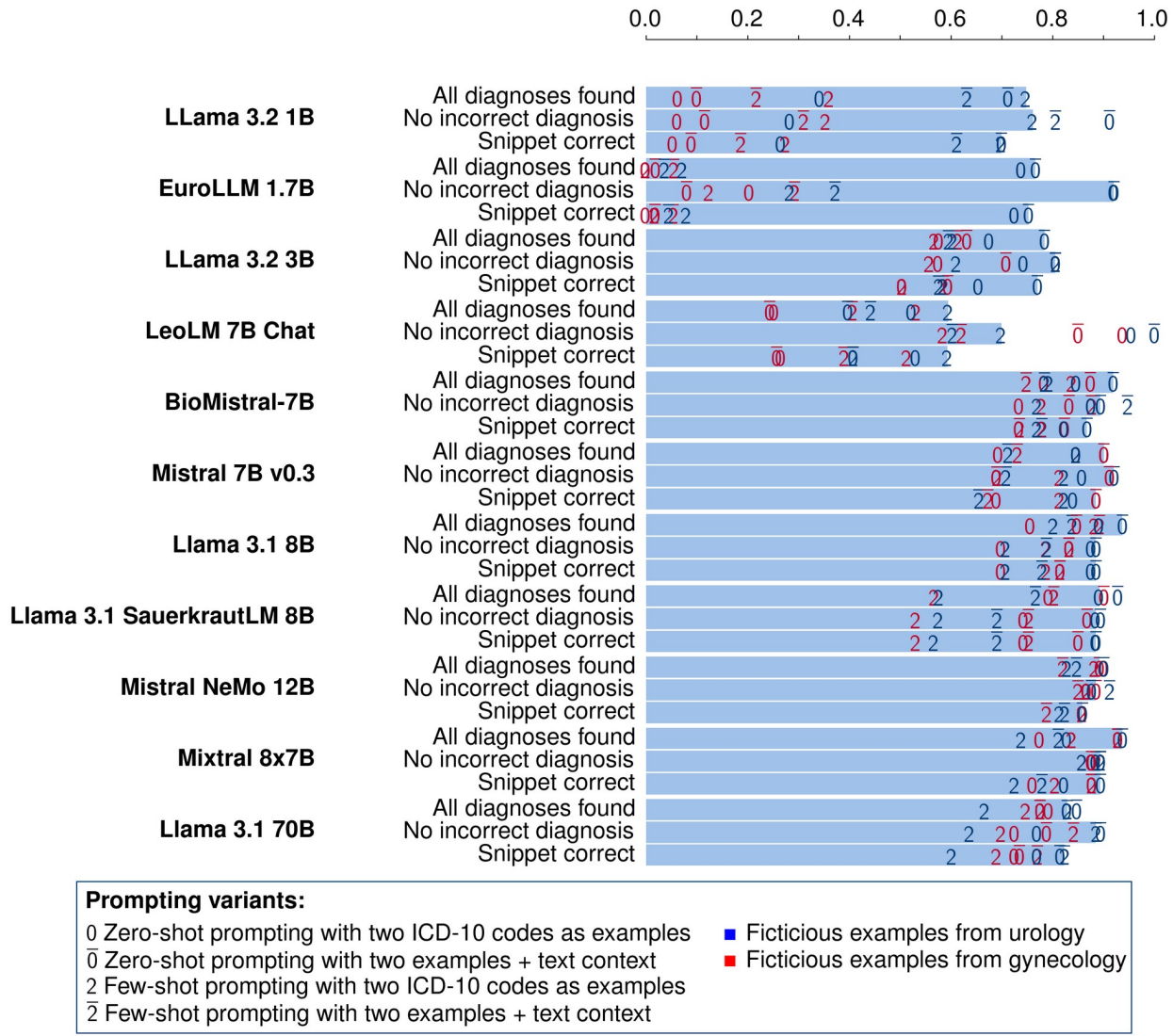

Figure 4 Proportion of correctly mapped ICD-10 codes using different models. The extracted diagnosis labels from the best model/prompt combination in Step 1 are used in the prompts for finding the ICD-10 codes. The values for different promping variants are indicated with the colored numbers. The bars indicate the best prompt with respect to the proportion of completely

- 17 -

*correct snippets. An interactive visualization of the data, designed to facilitate selective exploration, is available as Figure S2 in the supplementary materials.*

In Step 2 the performance of most models with at least 7 billion parameters is very close, with the exception of LeoLM, similar to Step 1. A clear advantage of the biggest models Mixtral 8x7B and Llama 3.1 70B cannot be seen here as their performance is very close to the models in the range from 7B to 12B. It is also interesting that BioMistral is not better than Mistral 7B in this task, although it has received more training with texts containing medical knowledge.

The question in Step 2 is very simple and zero-shot prompting is the best variant for almost all models. Adding the context of the whole snippet improves the results in most cases at least slightly.

The low performance of Llama 3.1 1B and EuroLLM 1.7B when using the fictitious examples from gynecology shows that they do not know the ICD-10 codes and entirely rely on the examples given in the prompt. This effect can also be seen in LeoLM, albeit less pronounced. For the other models, the difference between the examples from different domains is much smaller, which means that they are able to infer the ICD-10 codes using their built-in knowledge without help from the examples.

### Results for Step 3: Finding the first diagnosis date

Figure 5 shows the results for Step 3, i.e. the results for finding the first diagnosis dates for the ICD-10 coded tumor diagnoses extracted in Step 1 and Step 2. Same as in Step 2, we focus on the task-specific comparison of the models instead of the performance of the models employed on their own on all three documentation tasks. Accordingly, Figure 5 shows the results obtained using the results from the best models and prompts from the previous steps for constructing the prompts in Step 3. The detailed values for all the combinations can be found in Appendix A.

In addition to the LLMs, a rule-based approach is also included in Step 3 as a baseline. This approach uses the date in the same line as the occurrence of the diagnosis label that has been extracted via the best performing previous LLM output. In addition to taking the date from the same line, we also tried taking the closest date in terms of character distance with maximum line distance one or two. The achieved accuracies were 0.54 with the same line distance, 0.69 with a line distance of one, and 0.36 with a line distance of two.

The best results were achieved by Mistral NeMo 12B, which identified the first diagnosis date with an accuracy of 92% in the zero-shot approach. The ranking of the other models is similar to that in Step 1, with Llama 3.1 8B showing a performance similar to the bigger models, Mixtral 8x7B and Llama 3.1 70B. BioMistral, LeoLM 7B, and the smaller models performed worse than the simple heuristic approach, which uses the previously extracted diagnosis label and a regular expression for finding the closest date. There is no prompting variant that is clearly the best for all the models. The substantial variation introduced by different prompting strategies causes some models to outperform the heuristic baseline in certain configurations, but not in others.

This highlights the importance of prompt engineering and prompt evaluation for achieving optimal results in this context.

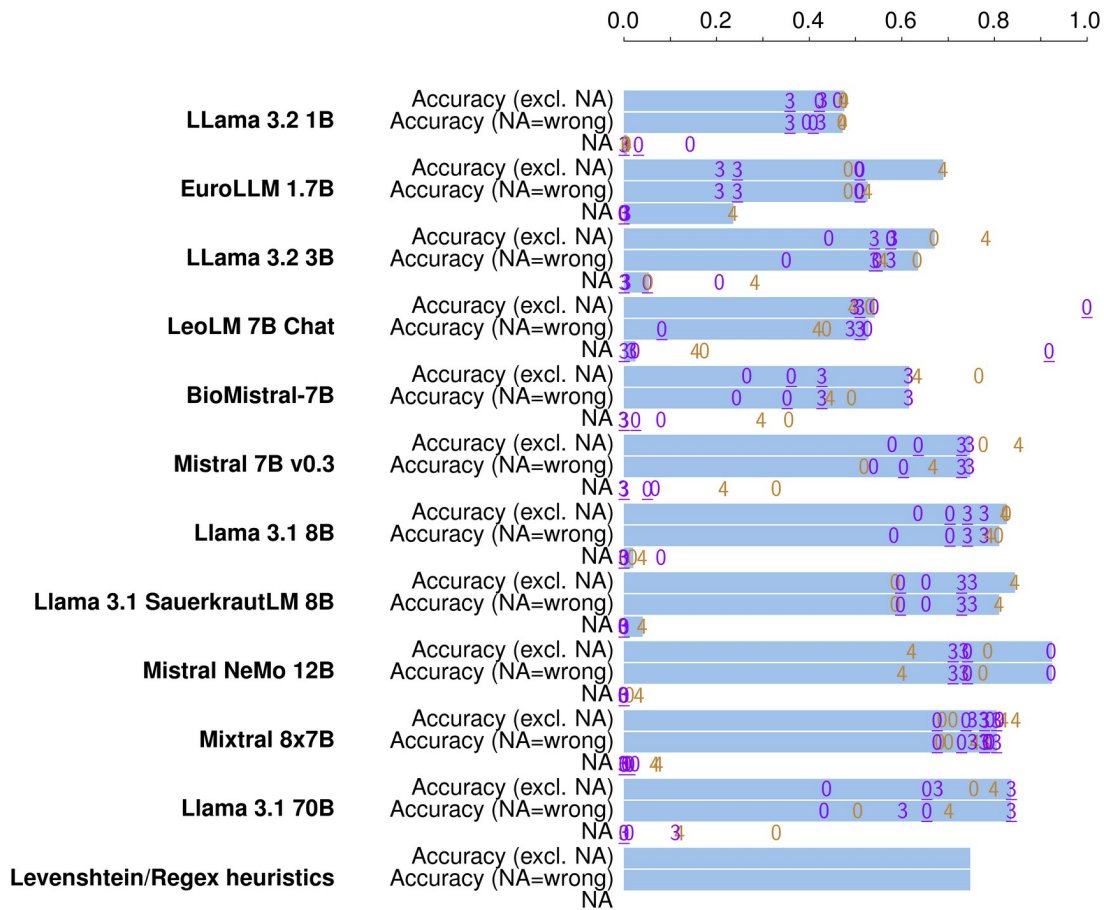

Figure 5 Proportion of correctly identified first diagnosis dates using different models. An interactive visualization of the data, designed to facilitate selective exploration, is available as Figure S3 in the supplementary materials.

In cases where a first diagnosis date had been annotated and the models predicted an incorrect date, the errors tended to be substantial. To quantify these errors, we calculated the difference in days between the predicted and correct dates, assuming the 15th of the month for month-only values and July 1st for year-only values. Based on this, the median error across all wrong LLM responses was 1,520 days (25th–75th percentile: 334 to 2,663 days).

# Discussion

## Limitations of the study and generalizability of results

In this study we aimed to highlight the capabilities of LLMs with respect to tumor documentation. A limitation concerning this is the fact that we only have doctors' notes from urology. We tried to address this by phrasing the core prompts in a general way without referring to things specific to urology. We then integrated information from either urology or a different medical domain into the examples that we added to the core prompts. By comparing the performance of symmetrical prompts with examples and instructions belonging to two different domains, we could investigate the impact of domain-specific instructions. The results therefore also allow some conclusions about the performance of the models and prompting techniques for tumor documentation in general.

Given the rather small size of our data set and its focus on a specific set of urology patients, the absolute numbers of the different metrics reported here might not reflect the exact overall performance of the models. We also could not cover all tasks that are part of the tumor documentation process. However, the comparison of different models and different prompting strategies for finding diagnoses and first diagnosis dates gives valuable insights into the behavior of the different models. This comparison can help identify which models are ideal starting points for fine-tuning towards information extraction for the purpose of tumor documentation.

The annotation of the tumor diagnoses in the text snippets used in this study was rather simple. The differences in the coding of the tumor diagnoses by the four different annotators were very small and mostly the result of isolated oversights, not uncertainty about how to interpret the content. The interpretation of the texts for deciding the attribution of a first diagnosis date had more shades of grey, however. This difficulty might be one reason why the results of many models in Step 3 are surprisingly not consistently better than the simple rule-based approach.

## Model performance and its connection to training data and model size

The results show that foundation models trained on a large amount of data from different domains like Llama 3.1 8B, Mistral 7B, and Mistral 12B are able to perform very specific tasks like tumor documentation. For performing the tasks presented here, the models need to understand the definition of a primary tumor diagnosis that is relevant for tumor documentation. Then they also need to connect the tumor diagnosis with the date of the initial diagnosis. With proper prompt design, they are able to grasp these abstract concepts from initial prompts that contain a task description. Short sequences of examples used in few-shot prompting further help their understanding.

We found that a plateau is quickly reached in few-shot prompting where most models do not profit from more examples. If the concepts in the prompt are known well-enough to the models zero-shot prompting is also sufficient, which seems to be the case with ICD-10 coding: Most models already had the best performance using zero-shot prompting in the ICD-10 coding task. More examples on an already clear concept seem to confuse the models, as the results here show. An exception to that could be seen in LeoLM 7B Chat and in the smaller models. In Step 1 and Step 2, LeoLM profited from more examples given in few-shot prompting than the other models. Presumably, this is the case because LeoLM saw much less training data in total compared to the other models with at least 7 billion parameters: LeoLM has received training with a dataset of 65 billion tokens in addition to the base model Llama 2 [48], which in turn has been trained on 2 trillion tokens [26]. In contrast to that, Llama 3.1 has been trained on more than 15 trillion tokens [49].

EuroLLM 1.7B was trained with 4 trillion tokens, with 50% of the training data being in English and 6% in German [30]. This means that EuroLLM has received more German training data than LeoLM and a similar amount of English training data. However, the results from LeoLM are clearly better by comparison. This may be due to the fact that a model with more parameters can hold the information from the training data better [50], e.g., remember the ICD-10 codes for the diseases, which are contained in the training data of all models as they are all trained on Wikipedia articles. Another factor may be that a certain model size is required for the models to perform well enough on such a complex task as the one presented here [50].

It has also been noted [51] that larger models are not always better in tasks with complex instructions. The resources needed for training and inference of larger models as well as the time and energy that is consumed by the models are also important points to consider. It is therefore important to find foundational models that are as small as possible but big enough to be trained and utilized for tumor documentation tasks. Our results point to models of 7-12 billion parameters as good candidates for this purpose.

The results also show that it is not so simple to improve upon foundation models by simply training them with some more data related to the area of expertise needed for the task. This can be seen in the results of the SauerkrautLM variant of Llama 3.1 8B, which does not improve upon the base model in this study, and in the results of BioMistral, which does not outperform Mistral 7B. However, numerous techniques are available to fine-tune such models, enhancing their performance on specific tasks substantially. In particular, there are techniques for parameter-efficient fine-tuning, which modify only a subset of the weights of the larger models [52]. A targeted fine-tuning of larger models seems more feasible and promising compared to a further training of foundational models with larger amounts of additional data that are broadly related to the topic in question.

## Contribution to openly available German clinical data

Especially in the medical domain, getting data for developing or evaluating methods is hard due to privacy protection constraints. There is only a small amount of datasets available but there are initiatives to publish more anonymized German clinical text documents [53,54].

The largest currently publicly available German dataset based on clinical documentation is the CARDIO:DE dataset, which comprises 500 discharge letters with 993,143 tokens from a cardiology department [55]. In the oncological domain, there is only one other publicly available German dataset containing anonymized real-world texts, which consists of shuffled sentences from 150 discharge summaries and has 89,942 tokens in total [56]. The number of letters in our

- 21 -This version of the article has been accepted for publication after peer review but is not the Version of Record and does not reflect post-acceptance improvements, or any corrections. The Version of Record is available online at: https://doi.org/10.1186/s13040-025-00463-8

data set is very similar (153) and the number of tokens in our data set is in the same order of magnitude (26,602 tokens using the Llama 3.1 tokenizer and 31,181 tokens using the tokenizer of Mistral 7B v0.3). Other data sets with German clinical texts contain texts translated from English [57] or synthetic data [58].

Therefore, the annotated data set of snippets from real-world doctors' notes that we release alongside this paper is in itself a valuable contribution to the field of clinical NLP with German data.

### Distinction to named entity recognition (NER) tasks and classification tasks

The objective of the tumor documentation task investigated here is to identify the tumor diagnoses associated with the patient. This makes the task different from named entity recognition (NER) tasks: NER could be used to find terms for tumors where they occur in the patient history. However, there are situations in which the occurrence of the term for a tumor in patient documentation does not imply a tumor diagnosis for the patient. This is, e.g., the case for suspected diagnoses, excluded diagnosis or tumor diagnoses from close relatives mentioned in the texts.

The task examined here can also not be solved via document-level classification. The first two steps can be addressed using multi-label classification but the step for identifying the initial diagnosis date, which relies on the extracted diagnosis text, cannot be solved this way.

As the evaluation scenario uses complex prompts to achieve the three different tasks, we did not include a comparison to BERT-based models [33] such as BioGottBERT [59] and medBERT.de [60], which were also trained on German medical data but with the purpose of NER tasks and text classification.

### ICD-10 coding

The BERT-based models BioGottBERT and medBERT.de have been evaluated on the ICD-10 classification on German discharge notes [60]. Precision values around 40% and recall (sensitivity) below 20% were reached in this study. Another comparison of the models Flan-T5 [61] and GermanBERT [62], which were trained on a set of 100,672 radiology reports, showed a maximum accuracy of 72.2% for predicting the correct three-character ICD-10 codes on the 50 most-used ICD-10 codes therein [63]. We achieved a higher performance of more than 85% accuracy for ICD-10 coding on our dataset. This performance is still not sufficient for most use cases. Yet, the fact that our results were achieved without any additional training or fine-tuning on the specific task could suggest that models of the size of 7-12 billion parameters can achieve a performance that is acceptable for real world use if trained properly.

ICD-10 coding for medical billing can be challenging even for big state-of-the-art LLMs such as GPT-4 [64]. It should be noted, however, that coding for medical billing differs substantially from the ICD-10 coding of tumors for the purpose of tumor documentation. In practice, these tasks are performed by completely different departments/people in the hospitals in Germany. The goal of coding for medical billing is to combine different ICD-10 codes for describing the patient condition in a way that leads to an optimal compensation from the cost bearer. Moreover, the number of relevant diagnoses for medical billing is, of course, much larger than the number of codes for tumor diagnoses in the ICD-10 catalogue. Therefore, coding for medical billing is a different and more difficult task than coding for tumor documentation. Indeed, this task is challenging even for humans: A study with a random sample of three departments of a German hospital found that only up to 56.7% of primary diagnoses were correctly coded there, and only up to 37.5% of secondary diagnoses [65].

LLMs such as GPT-4 have been evaluated on medical exams for students and they performed even better than many human professionals in these tests [66,67]. Therefore, even if the current performance of LLMs needs improvement on certain specialized tasks such as medical coding or tumor documentation, LLMs should be able to achieve at least a human-level performance in these areas in the future. The question remains how to achieve this goal. Optimal training techniques, the required training data, and the size of the models have to be determined for this purpose. There are also approaches for increasing the performance of LLMs on knowledge-intensive tasks which could be beneficial for ICD coding. A popular method for enhancing the knowledge of LLMs is retrieval augmented generation (RAG) [68], which has also been applied successfully in the biomedical domain in general [69,70] for improving the performance of ICD coding in particular [71].

## Conclusions

Based on a dataset of urological doctors' notes, this study aimed to evaluate the applicability of open-source large language models (LLMs) for tumor documentation in Germany. The results show that models such as Llama 3.1 8B, Mistral 7B, and Mistral NeMo 12B can be assigned the tasks of identifying tumor diagnoses, coding them with ICD-10, and finding the corresponding initial diagnosis dates. This performance was achieved through the use of carefully designed prompts, without any explicit training or fine-tuning of the models. The achieved accuracy values of around 90% for finding and coding diagnoses are not yet suitable for the real-world application of the models. A tailored fine-tuning of the models could, however, substantially improve the performance and make them viable tools for tumor documentation tasks.

We found that models of sizes in the range of 7-12 billion parameters offer a good balance of performance and resource efficiency. Larger models did not consistently outperform their smaller counterparts. The smaller models (Llama 3.2 1B and EuroLLM 1.7B) and the model trained with the least amount of data in total (LeoLM 7B Chat) struggled the most with the tasks. Llama 3.2 3B was not far behind the models of 7-12 billion parameters, and might also become a viable option if improved with other techniques that compensate for the smaller amount of knowledge contained in the model. Additional domain-related training or more training data in the target language does not always lead to an improved performance, as could be seen with BioMistral and Llama 3.1 SauerkrautLM. This indicates that the way of training has also a very big impact on the performance in addition to the size of the training data.

The results also underline the necessity of having independent validation datasets for evaluating the model performance on real clinical data. Together with the code for the evaluation, we release the anonymized and annotated dataset that served as the data basis for this study. This can serve as a valuable benchmark for the German clinical NLP community. Adding benchmarks from other clinical areas will be essential for enabling more comprehensive evaluation across use cases [72]. Our dataset can be seen as a first step toward developing a suite of benchmarks for tumor documentation in Germany, similar to what is already common in other areas of artificial intelligence research [73].

Our findings suggest that with proper training and prompt design, models in the 7-12 billion parameter range could be well-suited for the task of tumor documentation. However, further research is needed to enhance model reliability and performance for tumor documentation in practice. Future work could explore techniques like parameter-efficient fine-tuning or retrieval-augmented generation, which have already shown success in similar tasks [74]. With continued advancements in this area, LLMs could become powerful tools for supporting medical professionals in tumor documentation.

# Declarations

## Ethics approval and consent to participate

This study is based on a dataset of anonymized doctors' letters of patients that had passed away more than 10 years before the anonymization of the letters. The letters were collected and anonymized for a previous study. As there was no additional data acquisition or involvement of patients, an ethics approval was waived.

## Consent for publication

Not applicable. The manuscript does not contain data from any individual person.

## Availability of data and materials

The code for the evaluation is released under the MIT license and available from https://github.com/stefan-m-lenz/UroLlmEval. The data set used for this study is available from https://huggingface.co/datasets/stefan-m-lenz/UroLlmEvalSet.

## Competing interests

The authors declare no competing interests.

## Funding

This research has been supported by the Federal Ministry of Education and Research (BMBF) in Germany in the project "Digitaler FortschrittsHub Gesundheit – DECIDE" (FKZ 01ZZ2106A) of the German national medical informatics initiative. The project aims to facilitate data transfer and interoperability among different stakeholders in the healthcare sector and to provide decision support based on structured patient information.

## Authors' contributions

SL performed the text extraction from the PDFs, created the first annotation of the dataset, designed and conducted the experiments, and wrote the manuscript. AU was involved in method development and research and helped conducting experiments. MJ provided the anonymized data set and gave advice on legal aspects. MR contributed to the re-annotation and supported the integration of the re-annotation results in the manuscript with her experience as a medical doctor. TP supervised the work and provided the funding for the research. AU, MJ, MR, and TP further contributed by editing the manuscript. All authors read and approved the final version of the manuscript.

## Acknowledgements

We thank Lakisha Ortiz Rosario for supporting in various parts of the data analysis and in particular for implementing the plot design of the graphs in the paper and in the supplementary material. Furthermore, we thank Fatma Alickovic and Diana Perera Cutiño for participating in the re-annotation of the dataset.

# Appendix A: Detailed result tables

*Table A-1 Results of Step 1: Performance of different models for detecting a tumor diagnosis in a text. Unusable answers ("NA") count as wrong answers for calculating the accuracy here. These values are ignored when calculating sensitivity and specificity. The column headers indicate the prompt type: the number represents the count of examples, and the suffix specifies whether the examples come from a fictitious urology unit or a fictitious gynecological unit.*

|  |  | 0 | 2-gyn | 2-uro | 4-gyn | 4-uro | 6-gyn | 6-uro |
|---|---|---|---|---|---|---|---|---|
| BioMistral-7B | Accuracy (NA=wrong) | 0.49 | 0.77 | 0.75 | 0.70 | 0.79 | 0.58 | 0.64 |
|  | NA | 0.37 | 0.15 | 0.17 | 0.15 | 0.15 | 0.32 | 0.28 |
|  | Sensitivity | 1.00 | 0.96 | 0.91 | 0.84 | 0.93 | 0.87 | 0.91 |
|  | Specificity | 0.12 | 0.77 | 0.91 | 0.77 | 0.91 | 0.81 | 0.79 |
| EuroLLM 1.7B | Accuracy (NA=wrong) | 0.12 | 0.38 | 0.46 | 0.44 | 0.56 | 0.34 | 0.34 |
|  | NA | 0.79 | 0.03 | 0.04 | 0.03 | 0.07 | 0.03 | 0.01 |
|  | Sensitivity | 1.00 | 0.27 | 0.46 | 0.37 | 0.68 | 0.15 | 0.12 |
|  | Specificity | 0.00 | 0.80 | 0.53 | 0.69 | 0.38 | 0.97 | 1.00 |
| LLama 3.2 1B | Accuracy (NA=wrong) | 0.21 | 0.41 | 0.38 | 0.41 | 0.39 | 0.19 | 0.27 |
|  | NA | 0.14 | 0.26 | 0.15 | 0.17 | 0.18 | 0.27 | 0.09 |
|  | Sensitivity | 0.00 | 0.41 | 0.27 | 0.35 | 0.33 | 0.07 | 0.08 |
|  | Specificity | 0.97 | 0.86 | 0.89 | 0.86 | 0.85 | 0.96 | 0.89 |
| LLama 3.2 3B | Accuracy (NA=wrong) | 0.17 | 0.91 | 0.91 | 0.89 | 0.91 | 0.72 | 0.94 |
|  | NA | 0.81 | 0.00 | 0.03 | 0.00 | 0.01 | 0.00 | 0.00 |
|  | Sensitivity | 0.96 | 0.89 | 0.94 | 0.85 | 0.94 | 0.63 | 0.96 |
|  | Specificity | 0.80 | 0.97 | 0.92 | 1.00 | 0.83 | 1.00 | 0.89 |
| LeoLM 7B Chat | Accuracy (NA=wrong) | 0.38 | 0.58 | 0.62 | 0.59 | 0.60 | 0.58 | 0.70 |
|  | NA | 0.52 | 0.22 | 0.19 | 0.26 | 0.28 | 0.36 | 0.22 |
|  | Sensitivity | 1.00 | 0.99 | 0.93 | 1.00 | 1.00 | 0.99 | 1.00 |
|  | Specificity | 0.00 | 0.09 | 0.28 | 0.23 | 0.36 | 0.55 | 0.60 |
| Llama 3.1 70B | Accuracy (NA=wrong) | 0.35 | 0.83 | 0.86 | 0.83 | 0.87 | 0.85 | 0.88 |
|  | NA | 0.18 | 0.01 | 0.01 | 0.02 | 0.01 | 0.01 | 0.01 |
|  | Sensitivity | 0.23 | 0.82 | 0.88 | 0.85 | 0.88 | 0.88 | 0.89 |
|  | Specificity | 0.97 | 0.86 | 0.81 | 0.83 | 0.84 | 0.81 | 0.89 |
| Llama 3.1 8B | Accuracy (NA=wrong) | 0.74 | 0.91 | 0.96 | 0.93 | 0.98 | 0.90 | 0.98 |
|  | NA | 0.18 | 0.01 | 0.00 | 0.00 | 0.00 | 0.01 | 0.00 |
|  | Sensitivity | 1.00 | 0.91 | 0.98 | 0.91 | 1.00 | 0.89 | 0.99 |
|  | Specificity | 0.43 | 0.95 | 0.89 | 0.97 | 0.92 | 0.97 | 0.95 |
| Llama 3.1 SauerkrautLM 8B | Accuracy (NA=wrong) | 0.87 | 0.95 | 0.95 | 0.95 | 0.96 | 0.93 | 0.97 |
|  | NA | 0.06 | 0.02 | 0.01 | 0.01 | 0.01 | 0.01 | 0.02 |
|  | Sensitivity | 1.00 | 0.96 | 1.00 | 0.95 | 1.00 | 0.91 | 1.00 |
|  | Specificity | 0.73 | 0.97 | 0.83 | 0.97 | 0.89 | 1.00 | 0.94 |
| Mistral 7B v0.3 | Accuracy (NA=wrong) | 0.76 | 0.92 | 0.93 | 0.91 | 0.91 | 0.89 | 0.91 |
|  | NA | 0.11 | 0.00 | 0.00 | 0.03 | 0.00 | 0.05 | 0.00 |
|  | Sensitivity | 0.99 | 0.94 | 0.99 | 0.94 | 0.96 | 0.94 | 0.96 |
|  | Specificity | 0.31 | 0.86 | 0.73 | 0.89 | 0.76 | 0.92 | 0.76 |
| Mistral NeMo 12B | Accuracy (NA=wrong) | 0.82 | 0.93 | 0.93 | 0.93 | 0.93 | 0.93 | 0.94 |
|  | NA | 0.14 | 0.00 | 0.00 | 0.00 | 0.00 | 0.01 | 0.00 |
|  | Sensitivity | 0.97 | 1.00 | 1.00 | 0.95 | 0.99 | 0.94 | 1.00 |
|  | Specificity | 0.92 | 0.70 | 0.70 | 0.89 | 0.76 | 0.95 | 0.76 |
| Mixtral 8x7B | Accuracy (NA=wrong) | 0.81 | 0.94 | 0.92 | 0.89 | 0.91 | 0.84 | 0.93 |
|  | NA | 0.13 | 0.01 | 0.03 | 0.03 | 0.04 | 0.07 | 0.04 |
|  | Sensitivity | 0.92 | 0.95 | 0.96 | 0.91 | 0.96 | 0.87 | 0.97 |
|  | Specificity | 0.95 | 0.97 | 0.92 | 0.97 | 0.92 | 0.97 | 0.97 |

Table A-2 Results of Step 2: Proportion of snippets with correctly mapped ICD-10 codes for the different models and prompt types. The diagnosis labels from Step 1 are used in the prompts. The rows with only the names of the models in the first column use the results from the same model specified and the best prompt type for this model from Step 1. Additionally, the values using the best model/prompt combination overall from Step 1 are reported in the table. The column headers specify the prompt type: The number represents the number of examples, the suffix "uro"/"gyn" whether the prompt is tailed for urological diagnoses or diagnoses from a fictitious gynecological unit, and the suffix "ctx" indicates whether the full snippet has been added as context to the prompt.

|  | 0-uro | 0-gyn | 0-uro-ctx | 0-gyn-ctx | 2-uro | 2-gyn | 2-uro-ctx | 2-gyn-ctx |
|---|---|---|---|---|---|---|---|---|
| BioMistral-7B | 0.86 | 0.73 | 0.86 | 0.83 | 0.84 | 0.80 | 0.84 | 0.80 |
| BioMistral-7B - using results from best model in Step 1 | 0.82 | 0.73 | 0.87 | 0.82 | 0.77 | 0.78 | 0.78 | 0.73 |
| EuroLLM 1.7B | 0.52 | 0.02 | 0.57 | 0.01 | 0.18 | 0.03 | 0.08 | 0.07 |
| EuroLLM 1.7B - using results from best model in Step 1 | 0.73 | 0.00 | 0.75 | 0.02 | 0.08 | 0.02 | 0.04 | 0.05 |
| LLama 3.2 1B | 0.06 | 0.00 | 0.50 | 0.03 | 0.36 | 0.25 | 0.50 | 0.06 |
| LLama 3.2 1B - using results from best model in Step 1 | 0.27 | 0.05 | 0.70 | 0.09 | 0.70 | 0.27 | 0.61 | 0.19 |
| LLama 3.2 3B | 0.73 | 0.63 | 0.77 | 0.65 | 0.64 | 0.64 | 0.66 | 0.68 |
| LLama 3.2 3B - using results from best model in Step 1 | 0.65 | 0.50 | 0.77 | 0.59 | 0.58 | 0.50 | 0.58 | 0.58 |
| LeoLM 7B Chat | 0.54 | 0.39 | 0.43 | 0.28 | 0.62 | 0.60 | 0.48 | 0.53 |
| LeoLM 7B Chat - using results | 0.53 | 0.27 | 0.41 | 0.26 | 0.59 | 0.51 | 0.41 | 0.39 |

| | | | | | | | |
|---|---|---|---|---|---|---|---|
| from best model in Step 1 | | | | | | | |
| Llama 3.1 70B | 0.71 | 0.66 | 0.77 | 0.63 | 0.53 | 0.57 | 0.75 | 0.75 |
| Llama 3.1 70B - using results from best model in Step 1 | 0.77 | 0.73 | 0.81 | 0.73 | 0.60 | 0.69 | 0.82 | 0.77 |
| Llama 3.1 8B (the best model in step 1) | 0.88 | 0.70 | 0.88 | 0.81 | 0.71 | 0.79 | 0.78 | 0.81 |
| Llama 3.1 SauerkrautLM 8B | 0.88 | 0.75 | 0.88 | 0.87 | 0.60 | 0.58 | 0.70 | 0.79 |
| Llama 3.1 SauerkrautLM 8B - using results from best model in Step 1 | 0.88 | 0.74 | 0.88 | 0.85 | 0.57 | 0.53 | 0.69 | 0.75 |
| Mistral 7B v0.3 | 0.78 | 0.71 | 0.82 | 0.79 | 0.75 | 0.74 | 0.62 | 0.69 |
| Mistral 7B v0.3 - using results from best model in Step 1 | 0.84 | 0.69 | 0.88 | 0.88 | 0.82 | 0.81 | 0.65 | 0.67 |
| Mistral NeMo 12B | 0.86 | 0.85 | 0.84 | 0.83 | 0.79 | 0.82 | 0.81 | 0.80 |
| Mistral NeMo 12B - using results from best model in Step 1 | 0.86 | 0.86 | 0.86 | 0.86 | 0.81 | 0.86 | 0.82 | 0.79 |
| Mixtral 8x7B | 0.89 | 0.85 | 0.89 | 0.87 | 0.81 | 0.87 | 0.84 | 0.88 |
| Mixtral 8x7B - using results from best model in Step 1 | 0.82 | 0.76 | 0.89 | 0.88 | 0.73 | 0.81 | 0.78 | 0.88 |

Table A-3 Results of Step 3 for the different LLM models: Accuracy of finding the first diagnosis dates of the tumors. Unusable answers ("NA") count as wrong answers for calculating the accuracy here. The column headers indicate the prompt type: The number represents the number of examples, the suffix "dates" whether a selection of regex-extracted dates have been presented to the model in the prompt, and the suffix "verify" indicates the variant of Step 3 where the model is asked to verify single possible dates as first diagnosis dates.

| | 0 | 0-dates | 3 | 3-dates | 0-verify | 4-verify |
|---|---|---|---|---|---|---|
| BioMistral-7B | 0.20 | 0.37 | 0.68 | 0.44 | 0.41 | 0.43 |
| BioMistral-7B - using results from best models in previous steps | 0.25 | 0.35 | 0.62 | 0.43 | 0.49 | 0.45 |
| EuroLLM 1.7B | 0.52 | 0.52 | 0.17 | 0.38 | 0.51 | 0.36 |
| EuroLLM 1.7B - using results from best models in previous steps | 0.51 | 0.51 | 0.21 | 0.25 | 0.49 | 0.53 |
| LLama 3.2 1B | 0.46 | 0.42 | 0.41 | 0.42 | 0.41 | 0.41 |
| LLama 3.2 1B - using results from best models in previous steps | 0.40 | 0.41 | 0.43 | 0.36 | 0.47 | 0.47 |
| LLama 3.2 3B | 0.26 | 0.46 | 0.50 | 0.61 | 0.63 | 0.56 |
| LLama 3.2 3B - using results from best models in previous steps | 0.35 | 0.55 | 0.58 | 0.54 | 0.64 | 0.56 |
| LeoLM 7B Chat | 0.63 | 0.12 | 0.57 | 0.51 | 0.47 | 0.50 |
| LeoLM 7B Chat - using results from best models in previous steps | 0.53 | 0.08 | 0.49 | 0.51 | 0.44 | 0.42 |
| Llama 3.1 70B | 0.31 | 0.59 | 0.48 | 0.73 | 0.43 | 0.59 |
| Llama 3.1 70B - using results from best models in previous steps | 0.43 | 0.65 | 0.60 | 0.84 | 0.51 | 0.70 |
| Llama 3.1 8B | 0.59 | 0.71 | 0.78 | 0.74 | 0.81 | 0.79 |
| Llama 3.1 8B - using results from best models in previous steps | 0.58 | 0.70 | 0.78 | 0.74 | 0.81 | 0.79 |
| Llama 3.1 SauerkrautLM 8B | 0.61 | 0.56 | 0.73 | 0.71 | 0.58 | 0.79 |
| Llama 3.1 SauerkrautLM 8B - using results from best models in previous steps | 0.65 | 0.60 | 0.75 | 0.73 | 0.59 | 0.81 |
| Mistral 7B v0.3 | 0.53 | 0.58 | 0.72 | 0.71 | 0.51 | 0.63 |
| Mistral 7B v0.3 - using results from best models in previous steps | 0.54 | 0.60 | 0.75 | 0.73 | 0.52 | 0.67 |
| Mistral NeMo 12B | 0.88 | 0.73 | 0.74 | 0.72 | 0.77 | 0.63 |
| Mistral NeMo 12B - using results from best models in previous steps | 0.92 | 0.74 | 0.74 | 0.71 | 0.78 | 0.60 |
| Mixtral 8x7B (the best model in step 2) | 0.79 | 0.73 | 0.75 | 0.78 | 0.70 | 0.76 |

# Appendix B: Prompt templates (in German)

## Prompt templates for Step 1

| Prompt template "0" |
|---|
| **User:** Gibt es eine oder mehrere Tumordiagnosen in folgendem Diagnosetext? Als Tumordiagnose zählen hier Diagnosen, die im Kapitel II (Neubildungen/neoplasms) der International Classification of Diseases 10th revision (ICD-10) in den Kategorien C00-D48 beschrieben sind. Eine Tumorerkrankung in diesem Sinn ist eine Neubildung abnormen Gewebes aus körpereigenen Zellen im Körper des Patienten. Es ist entscheidend, nur Diagnosen als Tumordiagnosen zu werten, bei denen klar eine solche Tumorerkrankung beschrieben wird. Aussagen über Symptome, Therapien oder andere Erkrankungen sollten nicht als Tumordiagnosen interpretiert werden. Denken Sie nach, ob es sich bei den Diagnosen wirklich um Tumordiagnosen handelt und geben Sie nur Tumordiagnosen zurück, keine anderen Diagnosen. Antworten Sie mit einem JSON-Array, das die Tumordiagnosen als Strings beinhaltet oder mit einem leeren Array, falls es keine eindeutigen Tumordiagnosen gibt. Geben Sie im Antwort-Array keine Diagnosen an, die keine Tumordiagnosen sind. Gibt es eine oder mehrere Tumordiagnosen in folgendem Diagnosetext? Diagnosetext: [SNIPPET-TEXT] |

| Prompt template "2-gyn" |
|---|
| **User:** Gibt es eine oder mehrere Tumordiagnosen in folgendem Diagnosetext? Als Tumordiagnose zählen hier Diagnosen, die im Kapitel II (Neubildungen/neoplasms) der International Classification of Diseases 10th revision (ICD-10) in den Kategorien C00-D48 beschrieben sind. Eine Tumorerkrankung in diesem Sinn ist eine Neubildung abnormen Gewebes aus körpereigenen Zellen im Körper des Patienten. Es ist entscheidend, nur Diagnosen als Tumordiagnosen zu werten, bei denen klar eine solche Tumorerkrankung beschrieben wird. Aussagen über Symptome, Therapien oder andere Erkrankungen sollten nicht als Tumordiagnosen interpretiert werden. Denken Sie nach, ob es sich bei den Diagnosen wirklich um Tumordiagnosen handelt und geben Sie nur Tumordiagnosen zurück, keine anderen Diagnosen. Antworten Sie mit einem JSON-Array, das die Tumordiagnosen als Strings beinhaltet oder mit einem leeren Array, falls es keine eindeutigen Tumordiagnosen gibt. Geben Sie im Antwort-Array keine Diagnosen an, die keine Tumordiagnosen sind. <br> **Assistant:** Ich werde auf die folgenden Diagnosetexte kurz mit einem JSON-Array antworten |
| und nur Tumordiagnosen berücksichtigen. |
| **User:** Gibt es eine oder mehrere Tumordiagnosen in folgendem Diagnosetext?<br>Diagnosetext: 2005: Invasives lobuläres Mammakarzinom links, Stadium II<br>Sekundärbefunde: Verdacht auf axilläre Lymphknotenmetastasen links<br>BRCA1-Genmutation positiv |
| **Assistant:** ["Mammakarzinom"] |
| **User:** Gibt es eine oder mehrere Tumordiagnosen in folgendem Diagnosetext?<br>Diagnosetext: CA-125-Wert im Normbereich, unauffällige Sonographie. Kein Anhalt für ein Ovarialkarzinoms. |
| **Assistant:** [] (Der Verdacht auf eine erwähnte Tumorerkrankung, hier "Ovarialkarzinom", wird nicht bestätigt.) |
| **User:** Gibt es eine oder mehrere Tumordiagnosen in folgendem Diagnosetext?<br>Diagnosetext: [SNIPPET-TEXT] |
| **Prompt template "2-uro"** |
| **User:** Gibt es eine oder mehrere Tumordiagnosen in folgendem Diagnosetext? Als Tumordiagnose zählen hier Diagnosen, die im Kapitel II (Neubildungen/neoplasms) der International Classification of Diseases 10th revision (ICD-10) in den Kategorien C00-D48 beschrieben sind. Eine Tumorerkrankung in diesem Sinn ist eine Neubildung abnormen Gewebes aus körpereigenen Zellen im Körper des Patienten. Es ist entscheidend, nur Diagnosen als Tumordiagnosen zu werten, bei denen klar eine solche Tumorerkrankung beschrieben wird. Aussagen über Symptome, Therapien oder andere Erkrankungen sollten nicht als Tumordiagnosen interpretiert werden. Denken Sie nach, ob es sich bei den Diagnosen wirklich um Tumordiagnosen handelt und geben Sie nur Tumordiagnosen zurück, keine anderen Diagnosen. Antworten Sie mit einem JSON-Array, das die Tumordiagnosen als Strings beinhaltet oder mit einem leeren Array, falls es keine eindeutigen Tumordiagnosen gibt. Geben Sie im Antwort-Array keine Diagnosen an, die keine Tumordiagnosen sind. |
| **Assistant:** Ich werde auf die folgenden Diagnosetexte kurz mit einem JSON-Array antworten und nur Tumordiagnosen berücksichtigen. |
| **User:** Gibt es eine oder mehrere Tumordiagnosen in folgendem Diagnosetext?<br>Diagnosetext: Prostata-Karzinom pT1. Ausschluss Lebermetastasen |

**Assistant:** ["Prostata-Karzinom"]

**User:** Gibt es eine oder mehrere Tumordiagnosen in folgendem Diagnosetext?
Diagnosetext: Harnstauung links. Ausschluss eines Urothelkarzinoms

**Assistant:** [] (Das Urothelkarzinom wird ausgeschlossen.)

**User:** Gibt es eine oder mehrere Tumordiagnosen in folgendem Diagnosetext?
Diagnosetext: [SNIPPET-TEXT]

**Prompt template "4-gyn"**

**User:** Gibt es eine oder mehrere Tumordiagnosen in folgendem Diagnosetext? Als Tumordiagnose zählen hier Diagnosen, die im Kapitel II (Neubildungen/neoplasms) der International Classification of Diseases 10th revision (ICD-10) in den Kategorien C00-D48 beschrieben sind. Eine Tumorerkrankung in diesem Sinn ist eine Neubildung abnormen Gewebes aus körpereigenen Zellen im Körper des Patienten. Es ist entscheidend, nur Diagnosen als Tumordiagnosen zu werten, bei denen klar eine solche Tumorerkrankung beschrieben wird. Aussagen über Symptome, Therapien oder andere Erkrankungen sollten nicht als Tumordiagnosen interpretiert werden. Denken Sie nach, ob es sich bei den Diagnosen wirklich um Tumordiagnosen handelt und geben Sie nur Tumordiagnosen zurück, keine anderen Diagnosen. Antworten Sie mit einem JSON-Array, das die Tumordiagnosen als Strings beinhaltet oder mit einem leeren Array, falls es keine eindeutigen Tumordiagnosen gibt. Geben Sie im Antwort-Array keine Diagnosen an, die keine Tumordiagnosen sind.

**Assistant:** Ich werde auf die folgenden Diagnosetexte kurz mit einem JSON-Array antworten und nur Tumordiagnosen berücksichtigen.

**User:** Gibt es eine oder mehrere Tumordiagnosen in folgendem Diagnosetext?
Diagnosetext: 2005: Invasives lobuläres Mammakarzinom links, Stadium II
Sekundärbefunde: Verdacht auf axilläre Lymphknotenmetastasen links
BRCA1-Genmutation positiv

**Assistant:** ["Mammakarzinom"]

**User:** Gibt es eine oder mehrere Tumordiagnosen in folgendem Diagnosetext?
Diagnosetext: Die Patientin berichtet über tastbare Veränderung in der rechten Brust.
Die durchgeführte Mammographie und der Brustultraschall zeigen keine Anomalien,
und die Veränderung entspricht einer gutartigen zystischen Läsion.

| |
|---|
| MRT der Brust bestätigt das Fehlen maligner Merkmale.<br>Ausschluss eines Mammakarzinoms |
| **Assistant:** [] (Mammakarzinom wird ausgeschlossen) |
| **User:** Gibt es eine oder mehrere Tumordiagnosen in folgendem Diagnosetext?<br>Diagnosetext: CA-125-Wert im Normbereich, unauffällige Sonographie. Kein Anhalt für ein Ovarialkarzinoms. |
| **Assistant:** [] (Die Tumorerkrankung "Ovarialkarzinom" wird nicht bestätigt.) |
| **User:** Gibt es eine oder mehrere Tumordiagnosen in folgendem Diagnosetext?<br>Diagnosetext: Seröses Ovarialkarzinom des linken Ovars, keine Anzeichen für Fernmetastasen |
| **Assistant:** ["Ovarialkarzinom"] |
| **User:** Gibt es eine oder mehrere Tumordiagnosen in folgendem Diagnosetext?<br>Diagnosetext: [SNIPPET-TEXT] |

| **Prompt template "4-uro"** |
|---|
| **User:** Gibt es eine oder mehrere Tumordiagnosen in folgendem Diagnosetext? Als Tumordiagnose zählen hier Diagnosen, die im Kapitel II (Neubildungen/neoplasms) der International Classification of Diseases 10th revision (ICD-10) in den Kategorien C00-D48 beschrieben sind. Eine Tumorerkrankung in diesem Sinn ist eine Neubildung abnormen Gewebes aus körpereigenen Zellen im Körper des Patienten. Es ist entscheidend, nur Diagnosen als Tumordiagnosen zu werten, bei denen klar eine solche Tumorerkrankung beschrieben wird. Aussagen über Symptome, Therapien oder andere Erkrankungen sollten nicht als Tumordiagnosen interpretiert werden. Denken Sie nach, ob es sich bei den Diagnosen wirklich um Tumordiagnosen handelt und geben Sie nur Tumordiagnosen zurück, keine anderen Diagnosen. Antworten Sie mit einem JSON-Array, das die Tumordiagnosen als Strings beinhaltet oder mit einem leeren Array, falls es keine eindeutigen Tumordiagnosen gibt. Geben Sie im Antwort-Array keine Diagnosen an, die keine Tumordiagnosen sind. |
| **Assistant:** Ich werde auf die folgenden Diagnosetexte kurz mit einem JSON-Array antworten und nur Tumordiagnosen berücksichtigen. |
| **User:** Gibt es eine oder mehrere Tumordiagnosen in folgendem Diagnosetext? |
| Diagnosetext: Prostata-Karzinom pT1. Ausschluss Lebermetastasen |
| **Assistant:** ["Prostata-Karzinom"] |
| **User:** Gibt es eine oder mehrere Tumordiagnosen in folgendem Diagnosetext?<br>Diagnosetext: Miktionsbeschwerden. Ausschluss PCA. Z.n. Nierentransplantation 1980 links |
| **Assistant:** [] (Die Diagnose Prostatakarzinom/PCA ist hier ausgeschlossen.) |
| **User:** Gibt es eine oder mehrere Tumordiagnosen in folgendem Diagnosetext?<br>Diagnosetext: Harnstauung links. Ausschluss eines Urothelkarzinoms |
| **Assistant:** [] (Das Urothelkarzinom wird ausgeschlossen.) |
| **User:** Gibt es eine oder mehrere Tumordiagnosen in folgendem Diagnosetext?<br>Diagnosetext: Z.n. TUR-B-NR mit dem Nachweis eines Urothelkarzinoms der Harnblase pT2G3 |
| **Assistant:** ["Urothelkarzinom der Harnblase"] |
| **User:** Gibt es eine oder mehrere Tumordiagnosen in folgendem Diagnosetext?<br>Diagnosetext: [SNIPPET-TEXT] |

| **Prompt template "6-gyn"** |
|---|
| **User:** Gibt es eine oder mehrere Tumordiagnosen in folgendem Diagnosetext? Als Tumordiagnose zählen hier Diagnosen, die im Kapitel II (Neubildungen/neoplasms) der International Classification of Diseases 10th revision (ICD-10) in den Kategorien C00-D48 beschrieben sind. Eine Tumorerkrankung in diesem Sinn ist eine Neubildung abnormen Gewebes aus körpereigenen Zellen im Körper des Patienten. Es ist entscheidend, nur Diagnosen als Tumordiagnosen zu werten, bei denen klar eine solche Tumorerkrankung beschrieben wird. Aussagen über Symptome, Therapien oder andere Erkrankungen sollten nicht als Tumordiagnosen interpretiert werden. Denken Sie nach, ob es sich bei den Diagnosen wirklich um Tumordiagnosen handelt und geben Sie nur Tumordiagnosen zurück, keine anderen Diagnosen. Antworten Sie mit einem JSON-Array, das die Tumordiagnosen als Strings beinhaltet oder mit einem leeren Array, falls es keine eindeutigen Tumordiagnosen gibt. Geben Sie im Antwort-Array keine Diagnosen an, die keine Tumordiagnosen sind. |
| **Assistant:** Ich werde auf die folgenden Diagnosetexte kurz mit einem JSON-Array antworten |

und nur Tumordiagnosen berücksichtigen.

**User:** Gibt es eine oder mehrere Tumordiagnosen in folgendem Diagnosetext?
Diagnosetext: 2005: Invasives lobuläres Mammakarzinom links, Stadium II
Sekundärbefunde: Verdacht auf axilläre Lymphknotenmetastasen links
BRCA1-Genmutation positiv

**Assistant:** ["Mammakarzinom"]

**User:** Gibt es eine oder mehrere Tumordiagnosen in folgendem Diagnosetext?
Diagnosetext: Die Patientin berichtet über tastbare Veränderung in der rechten Brust.
Die durchgeführte Mammographie und der Brustultraschall zeigen keine Anomalien,
und die Veränderung entspricht einer gutartigen zystischen Läsion.
MRT der Brust bestätigt das Fehlen maligner Merkmale.
Ausschluss eines Mammakarzinoms

**Assistant:** [] (Mammakarzinom wird ausgeschlossen)

**User:** Gibt es eine oder mehrere Tumordiagnosen in folgendem Diagnosetext?
Diagnosetext: CA-125-Wert im Normbereich, unauffällige Sonographie. Kein Anhalt für ein Ovarialkarzinoms.

**Assistant:** [] (Die Tumorerkrankung "Ovarialkarzinom" wird nicht bestätigt.)

**User:** Gibt es eine oder mehrere Tumordiagnosen in folgendem Diagnosetext?
Diagnosetext: Seröses Ovarialkarzinom des linken Ovars, keine Anzeichen für Fernmetastasen

**Assistant:** ["Ovarialkarzinom"]

**User:** Gibt es eine oder mehrere Tumordiagnosen in folgendem Diagnosetext?
Diagnosetext: Ultraschall der Brust zeigt ein dichtes, aber homogenes Brustgewebe mit mehreren kleinen, nicht-suspekten Zysten.
Kein Anhalt für Malignität.
Diagnose: Fibrozystische Mastopathie.

**Assistant:** [] (Es wurde keine Tumorerkrankung bei der Patientin festgestellt.
Die hier gestellte Diagnose "Fibrozystische Mastopathie" (ICD-10-Code N60) zählt nicht als Tumordiagnose.)

**User:** Gibt es eine oder mehrere Tumordiagnosen in folgendem Diagnosetext?
Diagnosetext: Duktales In-situ-Karzinom der rechten Brust.
Keine Anzeichen für Fernmetastasen.
Hormonrezeptorstatus zeigt Östrogenrezeptor-Positivität. HER2/neu-Status negativ.

**Assistant:** ["Duktales In-situ-Karzinom der rechten Brust"]

**User:** Gibt es eine oder mehrere Tumordiagnosen in folgendem Diagnosetext?
Diagnosetext: [SNIPPET-TEXT]

### Prompt template "6-uro"

**User:** Gibt es eine oder mehrere Tumordiagnosen in folgendem Diagnosetext? Als Tumordiagnose zählen hier Diagnosen, die im Kapitel II (Neubildungen/neoplasms) der International Classification of Diseases 10th revision (ICD-10) in den Kategorien C00-D48 beschrieben sind. Eine Tumorerkrankung in diesem Sinn ist eine Neubildung abnormen Gewebes aus körpereigenen Zellen im Körper des Patienten. Es ist entscheidend, nur Diagnosen als Tumordiagnosen zu werten, bei denen klar eine solche Tumorerkrankung beschrieben wird. Aussagen über Symptome, Therapien oder andere Erkrankungen sollten nicht als Tumordiagnosen interpretiert werden. Denken Sie nach, ob es sich bei den Diagnosen wirklich um Tumordiagnosen handelt und geben Sie nur Tumordiagnosen zurück, keine anderen Diagnosen. Antworten Sie mit einem JSON-Array, das die Tumordiagnosen als Strings beinhaltet oder mit einem leeren Array, falls es keine eindeutigen Tumordiagnosen gibt. Geben Sie im Antwort-Array keine Diagnosen an, die keine Tumordiagnosen sind.

**Assistant:** Ich werde auf die folgenden Diagnosetexte kurz mit einem JSON-Array antworten und nur Tumordiagnosen berücksichtigen.

**User:** Gibt es eine oder mehrere Tumordiagnosen in folgendem Diagnosetext?
Diagnosetext: Prostata-Karzinom pT1. Ausschluss Lebermetastasen

**Assistant:** ["Prostata-Karzinom"]

**User:** Gibt es eine oder mehrere Tumordiagnosen in folgendem Diagnosetext?
Diagnosetext: Miktionsbeschwerden. Ausschluss PCA. Z.n. Nierentransplantation 1980 links

**Assistant:** [] (Die Diagnose Prostatakarzinom/PCA ist hier ausgeschlossen.)

**User:** Gibt es eine oder mehrere Tumordiagnosen in folgendem Diagnosetext?

| |
|---|
| Diagnosetext: Harnstauung links. Ausschluss eines Urothelkarzinoms |
| **Assistant:** [] (Das Urothelkarzinom wird ausgeschlossen.) |
| **User:** Gibt es eine oder mehrere Tumordiagnosen in folgendem Diagnosetext?<br>Diagnosetext: Z.n. TUR-B-NR mit dem Nachweis eines Urothelkarzinoms der Harnblase pT2G3 |
| **Assistant:** ["Urothelkarzinom der Harnblase"] |
| **User:** Gibt es eine oder mehrere Tumordiagnosen in folgendem Diagnosetext?<br>Diagnosetext: Digital-rektale Untersuchung zeigt eine vergrößerte, aber gleichmäßig geformte Prostata.<br>PSA-Wert im Normbereich. Diagnose: Benigne Prostatahyperplasie (BPH) ohne Hinweise auf Prostatakarzinom. |
| **Assistant:** [] (Der Verdacht auf die erwähnte Tumorerkrankung, hier "Prostatakarzinom", wird nicht bestätigt.<br>Die Diagnose "Benigne Prostatahyperplasie" (ICD-10-Code N40) zählt nicht als Tumordiagnose.) |
| **User:** Gibt es eine oder mehrere Tumordiagnosen in folgendem Diagnosetext?<br>Diagnosetext: Z.n. Adenomkarzinom der Prostata pT2b. ED 2015. |
| **Assistant:** ["Adenomkarzinom der Prostata"] |
| **User:** Gibt es eine oder mehrere Tumordiagnosen in folgendem Diagnosetext?<br>Diagnosetext: [SNIPPET-TEXT] |

## Prompt templates for Step 2

| |
|---|
| **Prompt template "0-gyn-ctx"** |
| **User:** [Diagnosetext]:<br>[SNIPPET-TEXT]<br>[Ende des Diagnosetexts]<br>Sie haben eine Diagnose im Diagnosetext identifiziert: "[DIAGNOSIS-LABEL]" Was ist der ICD- |

10-Code für diese Diagnose? Antworten Sie nur kurz mit dem 3-stelligen ICD-10-Code. Wenn es sich bei der Diagnose "[DIAGNOSIS-LABEL]" um ein Mammakarzinom handelt, antworten Sie zum Beispiel mit "C50", bei einem Ovarialkarzinom mit "C56". Die Metastasierung muss bei der Kodierung nicht berücksichtigt werden, das heißt die Codes C77, C78 und C79 für sekundäre Neubildungen sollten nicht verwendet werden. Die Antwort sollte also niemals "C77", "C78" oder "C79" sein.

**Prompt template "0-gyn"**

**User:** Was ist der ICD-10-Code für die Diagnose "[DIAGNOSIS-LABEL]"? Antworten Sie nur kurz mit dem 3-stelligen ICD-10-Code. Wenn es sich bei der Diagnose "[DIAGNOSIS-LABEL]" um ein Mammakarzinom handelt, antworten Sie zum Beispiel mit "C50", bei einem Ovarialkarzinom mit "C56". Die Metastasierung muss bei der Kodierung nicht berücksichtigt werden, das heißt die Codes C77, C78 und C79 für sekundäre Neubildungen sollten nicht verwendet werden. Die Antwort sollte also niemals "C77", "C78" oder "C79" sein.

**Prompt template "0-uro-ctx"**

**User:** [Diagnosetext]:
[SNIPPET-TEXT]
[Ende des Diagnosetexts]
Sie haben eine Diagnose im Diagnosetext identifiziert: "[DIAGNOSIS-LABEL]" Was ist der ICD-10-Code für diese Diagnose? Antworten Sie nur kurz mit dem 3-stelligen ICD-10-Code. Wenn es sich bei der Diagnose "[DIAGNOSIS-LABEL]" um ein Prostatakarzinom handelt, antworten Sie zum Beispiel mit "C61", bei einem Urothelkarzinom der Harnblase mit "C67". Die Metastasierung muss bei der Kodierung nicht berücksichtigt werden, das heißt die Codes C77, C78 und C79 für sekundäre Neubildungen sollten nicht verwendet werden. Die Antwort sollte also niemals "C77", "C78" oder "C79" sein.

**Prompt template "0-uro"**

**User:** Was ist der ICD-10-Code für die Diagnose "[DIAGNOSIS-LABEL]"? Antworten Sie nur kurz mit dem 3-stelligen ICD-10-Code. Wenn es sich bei der Diagnose "[DIAGNOSIS-LABEL]" um ein Prostatakarzinom handelt, antworten Sie zum Beispiel mit "C61", bei einem Urothelkarzinom der Harnblase mit "C67". Die Metastasierung muss bei der Kodierung nicht berücksichtigt werden, das heißt die Codes C77, C78 und C79 für sekundäre Neubildungen sollten nicht verwendet werden. Die Antwort sollte also niemals "C77", "C78" oder "C79" sein.

**Prompt template "2-gyn-ctx"**

**User:** Ihnen werden Diagnosetexte vorgelegt, zu denen Sie ICD-10-Codes zuordnen sollen. Die Frage ist jeweils: Was ist der ICD-10-Code für diese Diagnose? Antworten Sie darauf nur kurz mit dem 3-stelligen ICD-10-Code. Die Metastasierung muss bei der Kodierung nicht berücksichtigt werden, das heißt die Codes C77, C78 und C79 für sekundäre Neubildungen sollten nicht verwendet werden. Die Antwort sollte also niemals "C77", "C78" oder "C79" sein.

**Assistant:** Ich werde auf die folgenden Diagnosetexte kurz mit dem ICD-10-Code antworten.

**User:** Was ist der ICD-10-Code für diese Diagnose?
Diagnose: Metastasiertes Mammakarzinom

**Assistant:** C50

**User:** Was ist der 3-stellige ICD-10-Code für diese Diagnose?
Diagnose: Ovarialkarzinom

**Assistant:** C56

**User:** Was ist der 3-stellige ICD-10-Code für diese Diagnose?
Diagnose: "[DIAGNOSIS-LABEL]". Die Diagnose stammt aus folgendem längeren Text: [SNIPPET-TEXT]

**Prompt template "2-gyn"**

**User:** Ihnen werden Diagnosetexte vorgelegt, zu denen Sie ICD-10-Codes zuordnen sollen. Die Frage ist jeweils: Was ist der ICD-10-Code für diese Diagnose? Antworten Sie darauf nur kurz mit dem 3-stelligen ICD-10-Code. Die Metastasierung muss bei der Kodierung nicht berücksichtigt werden, das heißt die Codes C77, C78 und C79 für sekundäre Neubildungen sollten nicht verwendet werden. Die Antwort sollte also niemals "C77", "C78" oder "C79" sein.

**Assistant:** Ich werde auf die folgenden Diagnosetexte kurz mit dem ICD-10-Code antworten.

| |
|---|
| **User:** Was ist der ICD-10-Code für diese Diagnose?<br>Diagnose: Metastasiertes Mammakarzinom |
| **Assistant:** C50 |
| **User:** Was ist der 3-stellige ICD-10-Code für diese Diagnose?<br>Diagnose: Ovarialkarzinom |
| **Assistant:** C56 |
| **User:** Was ist der 3-stellige ICD-10-Code für diese Diagnose?<br>Diagnose: "[DIAGNOSIS-LABEL]". |

| **Prompt template "2-uro-ctx"** |
|---|
| **User:** Ihnen werden Diagnosetexte vorgelegt, zu denen Sie ICD-10-Codes zuordnen sollen. Die Frage ist jeweils: Was ist der ICD-10-Code für diese Diagnose? Antworten Sie darauf nur kurz mit dem 3-stelligen ICD-10-Code. Die Metastasierung muss bei der Kodierung nicht berücksichtigt werden, das heißt die Codes C77, C78 und C79 für sekundäre Neubildungen sollten nicht verwendet werden. Die Antwort sollte also niemals "C77", "C78" oder "C79" sein. |
| **Assistant:** Ich werde auf die folgenden Diagnosetexte kurz mit dem ICD-10-Code antworten. |
| **User:** Was ist der ICD-10-Code für diese Diagnose?<br>Diagnose: Metastasiertes Prostatakarzinom |
| **Assistant:** C61 |
| **User:** Was ist der 3-stellige ICD-10-Code für diese Diagnose?<br>Diagnose: Urothelkarzinom der Harnblase |
| **Assistant:** C67 |
| **User:** Was ist der 3-stellige ICD-10-Code für diese Diagnose?<br>Diagnose: "[DIAGNOSIS-LABEL]". Die Diagnose stammt aus folgendem längeren Text:<br>[SNIPPET-TEXT] |

| **Prompt template "2-uro"** |
|---|
| **User:** Ihnen werden Diagnosetexte vorgelegt, zu denen Sie ICD-10-Codes zuordnen sollen. Die Frage ist jeweils: Was ist der ICD-10-Code für diese Diagnose? Antworten Sie darauf nur kurz mit dem 3-stelligen ICD-10-Code. Die Metastasierung muss bei der Kodierung nicht berücksichtigt werden, das heißt die Codes C77, C78 und C79 für sekundäre Neubildungen sollten nicht verwendet werden. Die Antwort sollte also niemals "C77", "C78" oder "C79" sein. |
| **Assistant:** Ich werde auf die folgenden Diagnosetexte kurz mit dem ICD-10-Code antworten. |
| **User:** Was ist der ICD-10-Code für diese Diagnose?<br>Diagnose: Metastasiertes Prostatakarzinom |
| **Assistant:** C61 |
| **User:** Was ist der 3-stellige ICD-10-Code für diese Diagnose?<br>Diagnose: Urothelkarzinom der Harnblase |
| **Assistant:** C67 |
| **User:** Was ist der 3-stellige ICD-10-Code für diese Diagnose?<br>Diagnose: "[DIAGNOSIS-LABEL]". |

Prompt templates for Step 3

| **Prompt template "0-dates"** |
|---|
| **User:** [Diagnosetext]:<br>[SNIPPET-TEXT]<br>[Ende des Diagnosetexts]<br>Sie haben eine Diagnose im Diagnosetext identifiziert: "[DIAGNOSIS-LABEL]" Finden Sie eine Datumsangabe zur Erstdiagnose (ED) dieser Tumorerkankung im Text? Die Datumsangabe kann tagesgenau, monatsgenau oder nur eine Jahresangabe sein. Antworten Sie kurz mit dem Datum, wie es im Text angegeben ist. Achten Sie darauf, dass Sie nur ein Datum angeben, wenn es sich eindeutig auf die Erstdiagnose des Tumors bezieht. Datumsangaben |

zu Rezidiven, Metastasen oder zur Therapie des Tumors sind hier nicht relevant. Wenn es kein passendes Erstdiagnosedatum im Text gibt, antworten Sie mit "Nein".
Folgende Datumsangaben sind im Text zu finden: [DATES]. Ist eine der Datumsangaben das Datum der Erstdiagnose zur Tumordiagnose "[DIAGNOSIS-LABEL]"?

**Prompt template "0-verify"**

**User:** [Diagnosetext]:
[SNIPPET-TEXT]
[Ende des Diagnosetexts]
Im Diagnosetext haben Sie die Diagnose "[DIAGNOSIS-LABEL]" identifiziert. Zudem ist im Diagnosetext folgende Datumsangabe zu finden: 2002. Ist dies das Datum der Erstdiagnose der Tumorerkrankung "[DIAGNOSIS-LABEL]"? Antworten Sie mit "Ja" oder "Nein". Das Erstdiagnosedatum wird oft mit der Abkürzung "ED" vermerkt. Dann ist es klar, dass es sich um das Erstdiagnosedatum handelt. Es kann auch sein, dass das Datum ohne die Angabe "ED" angegeben ist, aber ein klarer Bezug zum ersten Diagnose der Tumorerkankung gegeben ist. In diesem Fall antworten Sie mit "Ja". Achten Sie aber darauf, dass Sie nur dann mit "Ja" antworten, wenn das Datum "[DATE]" sich eindeutig auf die Erstdiagnose zur Tumordiagnose "[DIAGNOSIS-LABEL]" bezieht. Datumsangaben, die sich ausschließlich auf Rezidive, Metastasen die Therapie des Tumors beziehen, sind hier nicht relevant. Wenn das Datum "[DATE]" sich nicht klar auf die Zeit (Tag, Monat oder Jahr) der Erstdiagnose bezieht, antworten Sie mit "Nein".

**Prompt template "0"**

**User:** [Diagnosetext]:
[SNIPPET-TEXT]
[Ende des Diagnosetexts]
Sie haben eine Diagnose im Diagnosetext identifiziert: "[DIAGNOSIS-LABEL]" Finden Sie eine Datumsangabe zur Erstdiagnose (ED) dieser Tumorerkankung im Text? Die Datumsangabe kann tagesgenau, monatsgenau oder nur eine Jahresangabe sein. Antworten Sie kurz mit dem Datum, wie es im Text angegeben ist. Achten Sie darauf, dass Sie nur ein Datum angeben, wenn es sich eindeutig auf die Erstdiagnose des Tumors bezieht. Datumsangaben zu Rezidiven, Metastasen oder zur Therapie des Tumors sind hier nicht relevant. Wenn es kein passendes Erstdiagnosedatum im Text gibt, antworten Sie mit "Nein".

| **Prompt template "3-dates"** |
|---|
| **User:** Finden Sie eine Datumsangabe zur Erstdiagnose (ED) einer gegebenen Tumorerkankung? Die Datumsangabe kann tagesgenau, monatsgenau oder nur eine Jahresangabe sein. Antworten Sie kurz mit dem Datum, wie es im Text angegeben ist. Achten Sie darauf, dass Sie nur ein Datum angeben, wenn es sich eindeutig auf die Erstdiagnose des Tumors bezieht. Datumsangaben zu Rezidiven, Metastasen oder zur Therapie des Tumors sind hier nicht relevant. Wenn es kein passendes Erstdiagnosedatum im Text gibt, antworten Sie mit "Nein". |
| **Assistant:** OK, geben Sie mir die Texte und Diagnosen. Ich werde mit dem Datum antworten, wie es im Text angegeben ist, wenn ich ein Erstdiagnosedatum zur Diagnose finde. Wenn es kein Datum dazu gibt, antworte ich mit "Nein". |
| **User:** [Diagnosetext]:<br>Z.n. Prostatakarzinom ED 1999<br>Z.n. Nierentransplantation 1980 links<br>[Ende des Diagnosetexts]<br>Im Diagnosetext haben Sie die Diagnose "Prostatakarzinom" identifiziert. Folgende Datumsangaben sind im Text zu finden: 1999, 1980. Ist eine der Datumsangaben das Datum der Erstdiagnose zur Tumordiagnose "Prostatakarzinom"? Antworten Sie mit dem Datum, wenn es die Zeit der Erstdiagnose der Tumordiagnose "Prostatakarzinom" beschreibt, oder mit "Nein". |
| **Assistant:** 1999 |
| **User:** [Diagnosetext]:<br>- TUR-B-NR mit dem Nachweis eines Urothelkarzinoms der Harnblase pT2G3<br>- Prostatakarzinom ED 07/2017<br>[Ende des Diagnosetexts]<br>Im Diagnosetext haben Sie die Diagnose "Urothelkarzinom" identifiziert. Im Diagnosetext ist folgende Datumsangabe zu finden: 07/2017. Ist dies das Datum der Erstdiagnose zur Tumordiagnose "Urothelkarzinom"? Antworten Sie mit dem Datum, wenn es die Zeit der Erstdiagnose der Tumordiagnose "Urothelkarzinom" beschreibt, oder mit "Nein". |
| **Assistant:** Nein (Es gibt kein Datum zum Urothelkarzinom, nur eines für das Prostatakarzinom) |
| **User:** [Diagnosetext]:<br>09/2011: Invasives lobuläres Mammakarzinom links, Stadium II |

| 10/2011: Brusterhaltende Therapie (BET) |
| 11/2011: Adjuvante Strahlentherapie, abgeschlossen 12/2011 |
| [Ende des Diagnosetexts] |
| Im Diagnosetext haben Sie die Diagnose "Mammakarzinom" identifiziert. Folgende Datumsangaben sind im Text zu finden: 09/2011, 10/2011, 11/2011, 12/2011. Ist eine der Datumsangaben das Datum der Erstdiagnose zur Tumordiagnose "Mammakarzinom"? Antworten Sie mit dem Datum, wenn es die Zeit der Erstdiagnose der Tumordiagnose "Mammakarzinom" beschreibt, oder mit "Nein". |

**Assistant:** 09/2011

**User:** [Diagnosetext]:
[SNIPPET-TEXT]
[Ende des Diagnosetexts]
Im Diagnosetext haben Sie die Diagnose "[DIAGNOSIS-LABEL]" identifiziert. Folgende Datumsangaben sind im Text zu finden: [DATES]. Ist eine der Datumsangaben das Datum der Erstdiagnose zur Tumordiagnose "[DIAGNOSIS-LABEL]"? Antworten Sie mit dem Datum, wenn es die Zeit der Erstdiagnose der Tumordiagnose "[DIAGNOSIS-LABEL]" beschreibt, oder mit "Nein".

| **Prompt template "3"** |

**User:** Finden Sie eine Datumsangabe zur Erstdiagnose (ED) einer gegebenen Tumorerkankung? Die Datumsangabe kann tagesgenau, monatsgenau oder nur eine Jahresangabe sein. Antworten Sie kurz mit dem Datum, wie es im Text angegeben ist. Achten Sie darauf, dass Sie nur ein Datum angeben, wenn es sich eindeutig auf die Erstdiagnose des Tumors bezieht. Datumsangaben zu Rezidiven, Metastasen oder zur Therapie des Tumors sind hier nicht relevant. Wenn es kein passendes Erstdiagnosedatum im Text gibt, antworten Sie mit "Nein".

**Assistant:** OK, geben Sie mir die Texte und Diagnosen. Ich werde mit dem Datum antworten, wie es im Text angegeben ist, wenn ich ein Erstdiagnosedatum zur Diagnose finde. Wenn es kein Datum dazu gibt, antworte ich mit "Nein".

**User:** Gibt es ein Erstdiagnosedatum zur Diagnose "Prostatakarzinom" in folgendem Text?
Text:
Z.n. Prostatakarzinom ED 1999
Z.n. Nierentransplantation 1980 links

| |
|---|
| **Assistant:** 1999 |
| **User:** Gibt es ein Erstdiagnosedatum zur Diagnose "Urothelkarzinom" in folgendem Text?<br>Text:<br>- TUR-B-NR mit dem Nachweis eines Urothelkarzinoms der Harnblase pT2G3<br>- Prostatakarzinom ED 07/2017 |
| **Assistant:** Nein (Es gibt kein Datum zum Urothelkarzinom, nur eines für das Prostatakarzinom.) |
| **User:** Gibt es ein Erstdiagnosedatum zur Diagnose "Mammakarzinom" in folgendem Text?<br>Text:<br>09/2011: Invasives lobuläres Mammakarzinom links, Stadium II<br>10/2011: Durchführung einer brusterhaltenden Therapie (BET)<br>11/2011: Beginn der adjuvanten Strahlentherapie, abgeschlossen 12/2011 |
| **Assistant:** 09/2011 |
| **User:** Gibt es ein Erstdiagnosedatum zur Diagnose "[DIAGNOSIS-LABEL]" in folgendem Text?<br>Text:<br>[SNIPPET-TEXT] |
| **Prompt template "4-verify"** |
| **User:** Sie erhalten Diagnosetexte und müssen bestimmen, ob ein bestimmtes Datum das Erstdiagnosedatum einer Tumorerkrankung ist. Antworten Sie mit "Ja" oder "Nein". Das Erstdiagnosedatum wird oft mit der Abkürzung "ED" vermerkt. Dann ist es klar, dass es sich um das Erstdiagnosedatum handelt. Es kann auch sein, dass das Datum ohne die Angabe "ED" angegeben ist, aber ein klarer Bezug zur ersten Diagnose der Tumorerkankung gegeben ist. In diesem Fall antworten Sie mit "Ja". Achten Sie aber darauf, dass Sie nur dann mit "Ja" antworten, wenn das genannte Datum sich eindeutig auf die Erstdiagnose zur Tumordiagnose bezieht. Datumsangaben, die sich ausschließlich auf Rezidive, Metastasen die Therapie des Tumors beziehen, sind hier nicht relevant. Wenn das Datum sich nicht klar auf die Zeit (Tag, Monat oder Jahr) der Erstdiagnose bezieht, antworten Sie mit "Nein". |
| **Assistant:** OK, geben Sie mir die Texte und die Datumsangaben. Ich werde mit "Ja" antworten, wenn das Datum die Erstdiagnosedatum zur Diagnose beschreibt, ansonsten antworte ich mit "Nein". |
| **User:** Ist "1999" die Datumsangabe für die Erstdiagnose der Tumorerkrankung "Prostatakarzinom" im Text? Text:<br>Z.n. Prostatakarzinom ED 1999<br>Z.n. Nierentransplantation 1980 links |
| **Assistant:** Ja. 1999 ist eindeutig als Erstdiagnosedatum (ED) bei der Diagnose "Prostatakarzinom" angegeben. |
| **User:** Ist "07/2017" die Datumsangabe für die Erstdiagnose der Tumorerkrankung "Urothelkarzinom" im Text? Text:<br>- TUR-B-NR mit dem Nachweis eines Urothelkarzinoms der Harnblase pT2G3<br>- Prostatakarzinom ED 07/2017 |
| **Assistant:** Nein. Es gibt kein Datum zum Urothelkarzinom, die Datumsangabe "07/2017" bezieht sich auf das Prostatakarzinom. |
| **User:** Ist "10/2011" die Datumsangabe für die Erstdiagnose der Tumorerkrankung "Mammakarzinonm" im Text? Text:<br>09/2011: Invasives lobuläres Mammakarzinom links, Stadium II<br>10/2011: Durchführung einer brusterhaltenden Therapie (BET)<br>11/2011: Beginn der adjuvanten Strahlentherapie, abgeschlossen 12/2011 |
| **Assistant:** Nein. Die Datumsangabe "10/2011" bezieht sich auf eine Therapie des Mammakarzinoms, nicht auf die Erstdiagnose der Tumorerkrankung. |
| **User:** Ist "09/2011" die Datumsangabe für die Erstdiagnose der Tumorerkrankung "Mammakarzinom" im Text? Text:<br>09/2011: Invasives lobuläres Mammakarzinom links, Stadium II<br>10/2011: Durchführung einer brusterhaltenden Therapie (BET)<br>11/2011: Beginn der adjuvanten Strahlentherapie, abgeschlossen 12/2011 |
| **Assistant:** Ja. Hier ist anzunehmen, dass sich die Datumsangabe auf die Erstdiagnose des Mammakarzinoms bezieht, auch wenn die Angabe nicht explizit mit "ED" gkennzeichnet ist. |
| **User:** Ist "[DATE]" die Datumsangabe für die Erstdiagnose der Tumorerkrankung "[DIAGNOSIS-LABEL]" im Text? Text:<br>[SNIPPET-TEXT] |